\documentclass{article}

\usepackage[main,preprint,nonatbib]{neurips_2026}

\usepackage[utf8]{inputenc} 
\usepackage[T1]{fontenc}    
\usepackage{hyperref}       
\usepackage{url}            
\usepackage{booktabs}       
\usepackage{amsfonts}       
\usepackage{nicefrac}       
\usepackage{microtype}      
\usepackage{xcolor}         

\usepackage[
    natbib=true,
    style=numeric,
    sorting=nyt,
    sortcites=true,
    maxbibnames=16,
    minbibnames=16,
    maxcitenames=1,
    mincitenames=1,
]{biblatex}
\AtBeginBibliography{\sloppy}

\usepackage{algorithm}
\usepackage{algorithmic}

\usepackage{times}
\usepackage{latexsym}

\usepackage{subcaption}
\usepackage{adjustbox}
\usepackage{amsmath}
\usepackage{stmaryrd}
\usepackage{mathtools}
\usepackage{amssymb}
\usepackage{cprotect}
\usepackage{cleveref}
\usepackage{dsfont}
\usepackage{todonotes}
\usepackage{tabularx}
\usepackage{multirow}
\usepackage{pgfplots}
\pgfplotsset{compat=1.18}
\usepackage{placeins}
\usepgfplotslibrary{groupplots}

\usepackage[scaled=0.90]{inconsolata}

\definecolor{linkcolor}{RGB}{99,148,237}
\hypersetup{
  colorlinks   = true, 
  urlcolor     = teal, 
  linkcolor    = linkcolor, 
  citecolor   = linkcolor, 
}

\newcommand{\Voc}{\mathcal{V}}

\newcommand{\sspace}{\hspace{10pt}}

\newcommand{\tokens}[1]{\texttt{"#1"}}
\newcommand{\intset}[2]{\{#1, \dots, #2\}}
\newcommand{\VNE}{\text{VNE}}
\newcommand{\KTC}{\text{KTC}}

\newcolumntype{C}{>{\centering\arraybackslash}X}
\newcolumntype{s}{>{\hsize=.3\hsize\linewidth=\hsize}C}

\usepackage{amsmath,amsfonts,bm}

\def\eqref#1{equation~\ref{#1}}

\def\1{\bm{1}}

\def\mA{{\bm{A}}}

\def\mD{{\bm{D}}}

\def\mI{{\bm{I}}}

\def\mK{{\bm{K}}}
\def\mL{{\bm{L}}}
\def\mM{{\bm{M}}}

\def\mU{{\bm{U}}}
\def\mV{{\bm{V}}}

\def\mmu{{\bm{\mu}}}
\def\mkappa{{\bm{\kappa}}}
\def\mPi{{\bm{\Pi}}}
\def\mLambda{{\bm{\Lambda}}}

\def\mSigma{{\bm{\Sigma}}}

\DeclareMathAlphabet{\mathsfit}{\encodingdefault}{\sfdefault}{m}{sl}
\SetMathAlphabet{\mathsfit}{bold}{\encodingdefault}{\sfdefault}{bx}{n}

\DeclareMathOperator{\Tr}{Tr}

\title{Kernel Token Contradiction: a Fast and Principled Approach for LLM Claim Uncertainty Quantification}

\author{%
  Jérémie Dentan\thanks{Equal contribution.}
  \sspace
  Alexi Canesse$^*$
  \sspace
  Mahammed El Sharkawy
  \sspace
  Sonia Vanier
\\[.8em]
LIX (École polytechnique, IP Paris, CNRS) \\
\texttt{jeremie.dentan@polytechnique.edu}\\
 \phantom{12}
\vspace{-.5cm}
}

\begin{document}

\maketitle
\setcounter{footnote}{0}
\begin{abstract}
Claim-level Uncertainty Quantification (UQ) aims to mitigate the lack of reliability of Large Language Models (LLMs) by evaluating the factuality of each claim in their outputs. We introduce Kernel Token Contradiction (KTC), a lightweight approach to compute claim-level UQ under realistic white-box conditions. KTC represents the candidate tokens involved in LLM generation as a positive semi-definite kernel that integrates both the LLM's conditional distribution and a token contradiction score. We then use the Von Neumann entropy to quantify the uncertainty of this kernel. To estimate token contradiction, we develop a new approach based on frequency statistics from the Wikipedia corpus. Although CPU-only, our approach achieves over an 8.2$\times$ speedup compared to state-of-the-art GPU-accelerated methods based on cross-encoders, and over a 65$\times$ speedup compared to CPU-only methods with comparable performance. Our evaluation spans two benchmarks across four European languages and 16 different models. KTC not only matches the average performance of existing methods but also outperforms them in high-precision regimes. This combination of computational efficiency and accuracy makes real-time monitoring of LLM outputs practical in production.
\end{abstract}
\section{Introduction} 

Despite their wide adoption, the reliability of Large Language Models (LLMs) remains an open challenge, as their outputs are prone to contain plausible yet non-factual content, known as hallucinations~\citep{huang_survey_2025, zhang_sirens_2025, sahoo_comprehensive_2024, ji_survey_2023}. Uncertainty Quantification (UQ) aims to provide LLM users with an estimate of the model's confidence, with the objective of mitigating the negative consequences of factuality hallucinations. In particular, claim-level UQ \citep{dentan_much_2026, fadeeva_factchecking_2024, vazquez_semeval2025_2025} provides a fine-grained view of the factuality of each claim in LLMs' outputs. This enables the user to be cautious about the uncertain parts without rejecting the whole answer, which is costly in terms of computation and user experience.

State-of-the-art claim-level UQ methods are known to have major limitations, highlighted by the MUCH benchmark \citep{dentan_much_2026}. First, their recall in high-precision regimes is insufficient. For instance, top-performing methods such as CCP \citep{fadeeva_factchecking_2024} only achieve an average precision of $69\%$ at recall $\le20\%$, which is largely insufficient to effectively mitigate hallucinations. Second, they rely on Natural Language Inference (NLI) models to estimate contradiction. These models not only require additional GPU resources, but they are also too slow for realistic monitoring of LLMs' outputs in production, representing up to over 100\% of the LLM generation runtime. Finally, these methods rely on NLI models trained on English corpora, which significantly alters their performance in other languages~\citep{dentan_much_2026}.

\begin{figure}[t!]
    \centering
    \includegraphics[width=\textwidth]{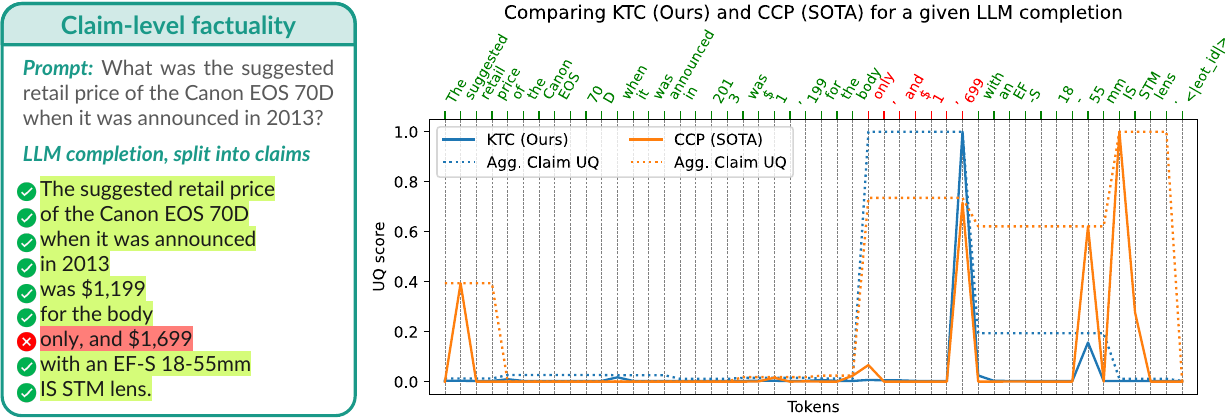}
    \caption{Example from the MUCH dataset~\citep{dentan_much_2026}: only one of the nine claims in the LLM generation is incorrect. KTC correctly identifies the wrong price (\$1{,}699), yielding a high uncertainty score for the incorrect claim, while CCP~\citep{fadeeva_factchecking_2024} produces several false positives and thus lower precision. Our method computes token-level scores that are aggregated at the claim level (see the dashed lines on the plot).}
    \label{fig:teaser}
\end{figure}

To mitigate these limitations, we developed a principled approach to detect contradictions between the candidate tokens emerging at each Next Token Prediction (NTP) step. Manual exploration revealed that these contradictions specifically arise in factuality hallucinations. For instance, when completing \tokens{and \$1,} with NTP (see \Cref{fig:teaser}), if the model computes a high probability for both tokens \tokens{699} and \tokens{499}, then it is uncertain about the price because it hesitates between two pieces of contradicting information. On the contrary, when completing \tokens{\$1,699 with}, if the model hesitates between \tokens{an} and \tokens{the}, these two possibilities do not contradict each other and the model simply hesitates between different formulations. Based on this intuition, we developed a new heuristic based on neighbor frequency statistics in the Wikipedia corpus to detect contradicting tokens. This heuristic is orders of magnitude faster than the NLI models used in top-performing methods such as SAR~\citep{duan_shifting_2024} or CCP~\citep{fadeeva_factchecking_2024}. Moreover, its setup is much simpler than training an NLI model, requiring only the tokenization of many Wikipedia pages, which only takes a few minutes of CPU computation.

\begin{figure}[t!]
    \centering
    \includegraphics[width=\textwidth]{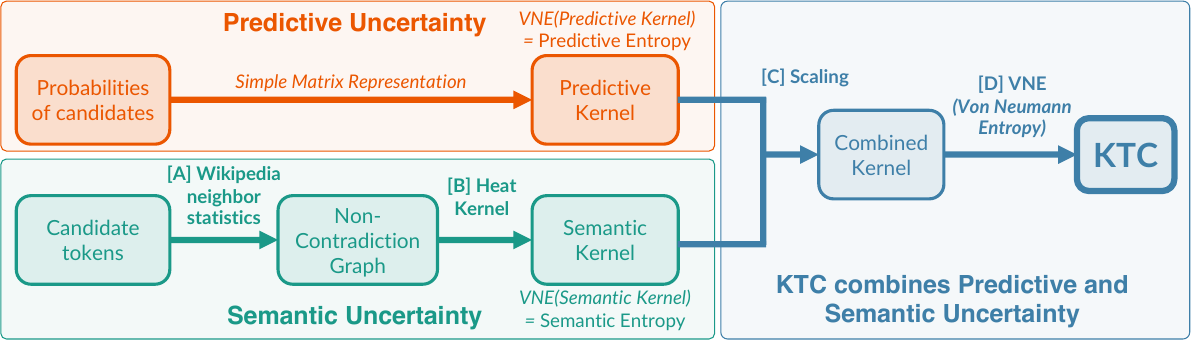}
    \caption{The KTC score is computed for each LLM-generated token. We construct a non-contradiction graph over candidate tokens (step [A]) and represent it as a PSD semantic kernel (step [B]). This kernel is then combined with the predictive kernel containing candidate probabilities (step [C]). Finally, we compute the Von Neumann entropy (VNE) of the combined kernel (step [D]).}
    \label{fig:pipeline}
\end{figure}

KTC consists of four steps, presented in \Cref{fig:pipeline}. It combines predictive uncertainty, derived from the entropy of the LLM output distribution, and semantic uncertainty, derived from contradictions between the meanings of candidate tokens. Steps [B] and [D] are inspired by Kernel Language Entropy (KLE)~\citep{nikitin_kernel_2024}, a generation-level UQ method that represents generation outcomes as PSD kernels and computes their Von Neumann Entropy (VNE). Compared to existing methods, our pipeline introduces three major methodological innovations. First, we adapt the principle of KLE from the generation level to the token level. Second, we replace the NLI model used in CCP or SAR with the neighbor frequency heuristic introduced above (step [A] in \Cref{fig:pipeline}). Finally, we scale semantic uncertainty by predictive uncertainty (step [C] in \Cref{fig:pipeline}). This scaling is absent in KLE but plays a crucial role in the performance of our approach at the token level (see \Cref{sec:scaling_predictive}).

We evaluated KTC on two benchmarks (MUCH~\citep{dentan_much_2026} and Mu-SHROOM~\citep{vazquez_semeval2025_2025}) spanning four languages (English, French, German, and Spanish) and 16 LLMs from various families. We compared KTC with comparable white-box claim-level UQ methods, including CCP~\citep{fadeeva_factchecking_2024} and SAR~\citep{duan_shifting_2024}. While CCP achieves the best overall performance, KTC ranks second with a small performance gap. Crucially, KTC is significantly more efficient, achieving at least an 8.2$\times$ speedup and requiring no GPU acceleration. Moreover, KTC clearly outperforms all competing methods in the high-precision regime, achieving an average precision of $78\%$ at recall $\le20\%$, compared to 69\% for CCP. The computational efficiency and high accuracy of KTC make it practical for real-time monitoring of LLMs outputs.

\textbf{Our contributions can be summarized as follows.}
\begin{itemize}
    \item We introduce a new heuristic based on token neighbor statistics to replace NLI models used in state-of-the-art claim-level UQ, with minimal computational and setup overhead;
    \item We propose a token-level score inspired by KLE~\citep{nikitin_kernel_2024}, integrating the proposed heuristic and a scaling by the LLM predictive distribution that is critical for token-level UQ;
    \item We evaluate KTC on benchmarks covering four European languages and 16 LLMs, achieving comparable average performance to state-of-the-art methods while being substantially faster, and largely outperforming all methods in the high-precision regime;
    \item We release all source code and hyperparameters needed to reproduce our experiments.\footnote{See \url{https://github.com/orailix/kernel_token_contradiction} and \Cref{app:reproducible_science}.}
\end{itemize}

\section{Background} \label{sec:background}

We briefly recall the uncertainty quantification tools and kernel methods principles underlying KTC.

\paragraph{Shannon's Entropy~\citep{shannon_mathematical_1948}}
Let $\delta \in \mathbb{N}^*$ and $p = \{p_i\}_{1 \le i \le \delta} \in [0, 1]^\delta$ such that $\sum p_i = 1$ be a probability distribution over a finite set. Given a random variable $X \sim p$, Shannon's Entropy quantifies the degree of uncertainty of the system. It is defined as $H(X) = - \sum\nolimits_{i=1}^{\delta} p_i \log{p_i}$, with the limit $x\log{x} \to 0$ as $x\to 0_+$. For instance, if $p_1 = 1$ and thus $p_{i>1} = 0$, the entropy is minimal and equal to zero; conversely, it is maximal and equal to $\log{\delta}$ when the $p_i$'s are all equal to $1/\delta$.

\paragraph{Von Neumann Entropy (VNE)~\citep{vonneumann_mathematical_1955}} Let $\rho \in \mathbb{R}^{\delta\times \delta}$ be a density matrix describing a quantum system, meaning that $\rho$ is positive semi-definite and $\Tr(\rho) = 1$. Let $\lambda_1, \dots, \lambda_\delta$ be the eigenvalues of $\rho$. VNE quantifies the quantum uncertainty of the system described by $\rho$. It is defined as

\begin{equation}
    \VNE = -\Tr(\rho \log \rho) = -\sum_{i=1}^{\delta}\lambda_i\log \lambda_i.
\end{equation}

\paragraph{Predictive Entropy}
Let $\Voc = \{t_1, \dots, t_N\}$ be a vocabulary consisting of $N$ distinct tokens, and let $f: \Voc^* \rightarrow [0,1]^N$ denote an LLM. It is trained using the Next Token Prediction (NTP) objective to approximate a target language modeling distribution $P$. For a given prefix sequence $x \in \Voc^*$, the $i$-th component of $f(x)$ estimates the probability that the \textit{candidate token} $t_i$ follows the prefix. Therefore, as training progresses, we aim to have the convergence $[f(x)]_i \longrightarrow P(t_i \mid x)$.

Let $\delta \in \mathbb{N}^*$ denote a maximum number of candidate tokens. For $i \in \intset{1}{\delta}$, let $\Delta^\delta_i(x) \in \intset{1}{N}$ denote the index of the $i$-th most likely candidate with respect to $f(x)$. Let $q_i \coloneqq [f(x)]_{\Delta^\delta_i(x)}$ be the conditional probability of the $i$-th most likely candidate. We define the \textit{predictive entropy} as Shannon's entropy of this discrete probability distribution. This quantity is also known as the \textit{token entropy} in the literature~\citep{dentan_much_2026,guerreiro_looking_2023} and defined as

\begin{equation}
    \text{Predictive Entropy} = -\sum_{i = 1}^{\delta} \frac{q_i}{\sum_{1\le k \le \delta} q_k} \log{\frac{q_i}{\sum_{1\le k \le \delta} q_k}}.
\end{equation}

\paragraph{Epistemic and Aleatoric Uncertainty} 

A common distinction is made between \textit{aleatoric uncertainty}, which captures the inherent randomness of LLM generations, and \textit{epistemic uncertainty}, which arises from a lack of knowledge required to make accurate predictions~\citep{hullermeier_aleatoric_2021}. In the context of LLMs, aleatoric uncertainty is desirable to ensure that the model remains creative, whereas epistemic uncertainty is associated with factuality hallucinations~\citep{yadkori_believe_2024}. However, predictive entropy does not account for the \textit{semantics} of tokens and therefore only captures the entropy of the \textit{predictive distribution} over candidate tokens, without distinguishing between epistemic and aleatoric uncertainty. Drawing inspiration from previous works~\citep{kuhn_semantic_2023, farquhar_detecting_2024, nikitin_kernel_2024}, we propose a method that estimates the entropy of this distribution while accounting for semantic relations between tokens to isolate epistemic uncertainty.
\section{Kernel Token Contradiction} \label{sec:methodology}

In this section we introduce KTC, our new contradiction-based approach for computing claim-level uncertainty in LLM outputs. Our method consists of four stages, illustrated in \Cref{fig:pipeline}. For clarity, steps [B] and [D] are presented together in \Cref{sec:steps_b_d}, while step [C] is presented in \Cref{sec:scaling_predictive}.

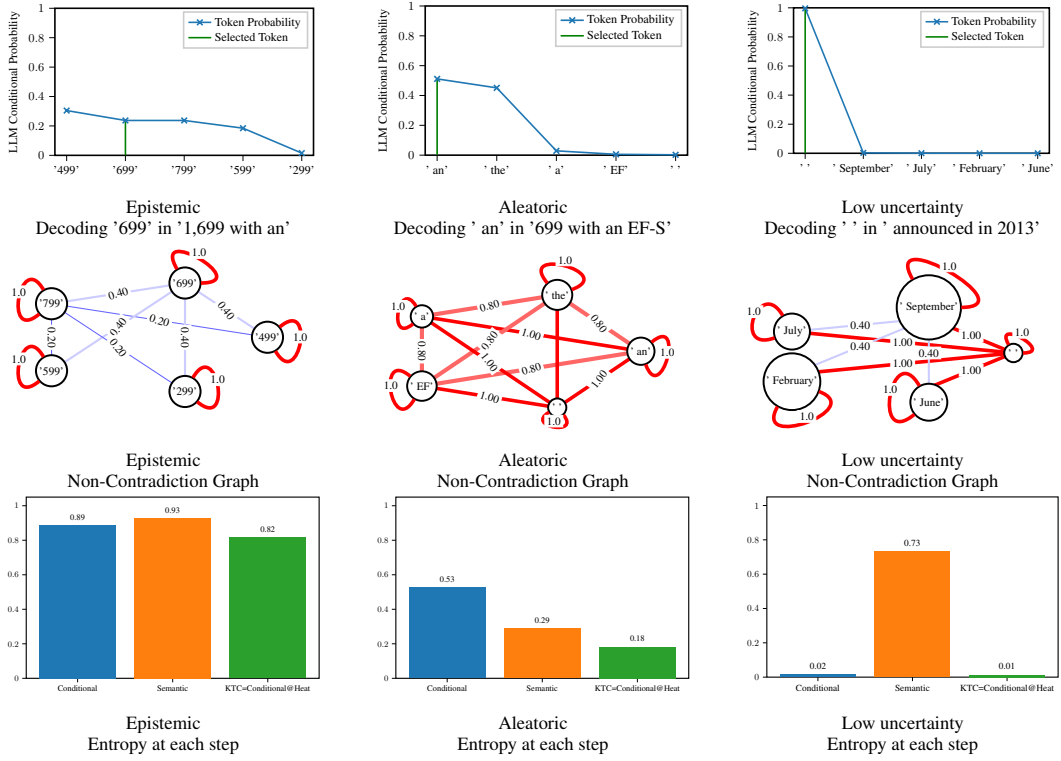
\begin{figure}[h!]
	\centering
	\setlength{\abovecaptionskip}{7pt}
	\begin{subfigure}{.32\textwidth}
		\centering
		\small
		\resizebox{\textwidth}{!}{
\begin{tikzpicture}

	\definecolor{darkgray176}{RGB}{176,176,176}
	\definecolor{green}{RGB}{0,128,0}
	\definecolor{lightgray204}{RGB}{204,204,204}
	\definecolor{steelblue31119180}{RGB}{31,119,180}

	\begin{groupplot}[group style={group size=3 by 3}]
		\nextgroupplot[
			height=1.4\textwidth,
			width=1.6\textwidth,
			every axis plot/.append style={line width=2pt},
			every mark/.append style={solid}, very thick,
			legend cell align={left},
			legend style={fill opacity=0.8, draw opacity=1, text opacity=1, draw=lightgray204, font=\normalsize},
			tick align=outside,
			tick pos=left,
			x grid style={darkgray176},
			xmin=-0.2, xmax=4.2,
			xtick style={color=black},
			xtick={0,1,2,3,4},
			xticklabels={\tokens{499},\tokens{699},\tokens{799},\tokens{599},\tokens{299}},
			y grid style={darkgray176},
			label style={font=\large},
			tick label style={font=\normalsize},
			ylabel={LLM Conditional Probability},
			ymin=0, ymax=1,
			ytick style={color=black}
		]
		\addplot [very thick, steelblue31119180, mark=x, mark size=4, mark options={solid}]
		table {%
				0 0.30452573299408
				1 0.237164884805679
				2 0.237164884805679
				3 0.184704199433327
				4 0.0151614435017109
			};
		\addlegendentry{Token Probability}
		\addplot [very thick, green]
		table {%
				1 0
				1 0.237164884805679
			};
		\addlegendentry{Selected Token}
	\end{groupplot}

\end{tikzpicture}}
		\caption*{\centering\small \textbf{Epistemic}\\ {\scriptsize Decoding \tokens{699} in \tokens{1,699 with an}}}
	\end{subfigure}
	\hfill
	\begin{subfigure}{.315\textwidth}
		\centering
		\small
		\resizebox{\textwidth}{!}{
\begin{tikzpicture}

	\definecolor{darkgray176}{RGB}{176,176,176}
	\definecolor{green}{RGB}{0,128,0}
	\definecolor{lightgray204}{RGB}{204,204,204}
	\definecolor{steelblue31119180}{RGB}{31,119,180}

	\begin{groupplot}[group style={group size=3 by 3}]
		\nextgroupplot[
			height=1.4\textwidth,
			width=1.6\textwidth,
			every axis plot/.append style={line width=2pt},
			every mark/.append style={solid}, very thick,
			legend cell align={left},
			legend style={fill opacity=0.8, draw opacity=1, text opacity=1, draw=lightgray204, font=\normalsize},
			tick align=outside,
			tick pos=left,
			x grid style={darkgray176},
			xmin=-0.2, xmax=4.2,
			xtick style={color=black},
			xtick={0,1,2,3,4},
			xticklabels={\tokens{ an},\tokens{ the},\tokens{ a},\tokens{ EF},\tokens{ }},
			y grid style={darkgray176},
			label style={font=\large},
			tick label style={font=\normalsize},
			ylabel={LLM Conditional Probability},
			ymin=0, ymax=1,
			ytick style={color=black}
		]
		\addplot [very thick, steelblue31119180, mark=x, mark size=4, mark options={solid}]
		table {%
				0 0.51124507188797
				1 0.451172202825546
				2 0.0288424752652645
				3 0.00567942019551992
				4 0.00236753467470407
			};
		\addlegendentry{Token Probability}
		\addplot [very thick, green]
		table {%
				0 0
				0 0.51124507188797
			};
		\addlegendentry{Selected Token}
	\end{groupplot}

\end{tikzpicture}}
		\caption*{\centering\small \textbf{Aleatoric}\\ {\scriptsize Decoding \tokens{ an} in \tokens{699 with an EF-S}}}
	\end{subfigure}
	\hfill
	\begin{subfigure}{.32\textwidth}
		\centering
		\small
		\resizebox{\textwidth}{!}{
\begin{tikzpicture}

	\definecolor{darkgray176}{RGB}{176,176,176}
	\definecolor{green}{RGB}{0,128,0}
	\definecolor{lightgray204}{RGB}{204,204,204}
	\definecolor{steelblue31119180}{RGB}{31,119,180}

	\begin{groupplot}[group style={group size=3 by 3}]
		\nextgroupplot[
			height=1.4\textwidth,
			width=1.6\textwidth,
			every axis plot/.append style={line width=2pt},
			every mark/.append style={solid}, very thick,
			legend cell align={left},
			legend style={fill opacity=0.8, draw opacity=1, text opacity=1, draw=lightgray204, font=\normalsize},
			tick align=outside,
			tick pos=left,
			x grid style={darkgray176},
			xmin=-0.2, xmax=4.2,
			xtick style={color=black},
			xtick={0,1,2,3,4},
			xticklabels={\tokens{ },\tokens{ September},\tokens{ July},\tokens{ February},\tokens{ June}},
			y grid style={darkgray176},
			label style={font=\large},
			tick label style={font=\normalsize},
			ylabel={LLM Conditional Probability},
			ymin=0, ymax=1,
			ytick style={color=black}
		]
		\addplot [very thick, steelblue31119180, mark=x, mark size=4, mark options={solid}]
		table {%
				0 0.996218383312225
				1 0.00279817264527082
				2 0.000707488739863038
				3 6.58067074255086e-05
				4 5.80742162128445e-05
			};
		\addlegendentry{Token Probability}
		\addplot [very thick, green]
		table {%
				0 0
				0 0.996218383312225
			};
		\addlegendentry{Selected Token}
	\end{groupplot}

\end{tikzpicture}}
		\caption*{\centering\small \textbf{Low uncertainty}\\ {\scriptsize Decoding \tokens{ } in \tokens{ announced in 2013}}}
	\end{subfigure}
	\\[1.6em]
	\begin{subfigure}{.32\textwidth}
		\centering
		\small
		\resizebox{\textwidth}{!}{
\begin{tikzpicture}

	\definecolor{darkgray176}{RGB}{176,176,176}
	\definecolor{lavender204204255}{RGB}{204,204,255}
	\definecolor{mediumslateblue102102255}{RGB}{102,102,255}

	\begin{groupplot}[group style={group size=3 by 3}]
		\nextgroupplot[
		height=1.4\textwidth,
		width=1.6\textwidth,
		every axis plot/.append style={line width=2pt},
		every mark/.append style={solid}, very thick,
		hide x axis,
		hide y axis,
    font=\LARGE,
		scaled x ticks=manual:{}{\pgfmathparse{#1}},
		scaled y ticks=manual:{}{\pgfmathparse{#1}},
		tick align=outside,
		x grid style={darkgray176},
		xmajorticks=false,
		xmin=-1.198963847427483, xmax=1.48994679093202,
		xtick style={color=black},
		xticklabels={},
		y grid style={darkgray176},
		ymajorticks=false,
		ymin=-1.653366454046, ymax=1.65063937800048,
		ytick style={color=black},
		yticklabels={}
		]
		\path [draw=red, line width=2.75pt]
		(axis cs:1,0)
		--(axis cs:1,0);

		\path [draw=lavender204204255, line width=1.6pt]
		(axis cs:1,0)
		--(axis cs:0.309016952187328,0.951056568356054);

		\path [draw=mediumslateblue102102255, line width=0.8pt]
		(axis cs:1,0)
		--(axis cs:-0.70901705649546,0.587785261505403);

		\path [draw=red, line width=2.75pt]
		(axis cs:0.309016952187328,0.951056568356054)
		--(axis cs:0.309016952187328,0.951056568356054);

		\path [draw=lavender204204255, line width=1.6pt]
		(axis cs:0.309016952187328,0.951056568356054)
		--(axis cs:-0.70901705649546,0.587785261505403);

		\path [draw=lavender204204255, line width=1.6pt]
		(axis cs:0.309016952187328,0.951056568356054)
		--(axis cs:-0.709016996890813,-0.58778532111005);

		\path [draw=lavender204204255, line width=1.6pt]
		(axis cs:0.309016952187328,0.951056568356054)
		--(axis cs:0.309017101198944,-0.951056508751408);

		\path [draw=red, line width=2.75pt]
		(axis cs:-0.70901705649546,0.587785261505403)
		--(axis cs:-0.70901705649546,0.587785261505403);

		\path [draw=mediumslateblue102102255, line width=0.8pt]
		(axis cs:-0.70901705649546,0.587785261505403)
		--(axis cs:-0.709016996890813,-0.58778532111005);

		\path [draw=mediumslateblue102102255, line width=0.8pt]
		(axis cs:-0.70901705649546,0.587785261505403)
		--(axis cs:0.309017101198944,-0.951056508751408);

		\path [draw=red, line width=2.75pt]
		(axis cs:-0.709016996890813,-0.58778532111005)
		--(axis cs:-0.709016996890813,-0.58778532111005);

		\path [draw=red, line width=2.75pt]
		(axis cs:0.309017101198944,-0.951056508751408)
		--(axis cs:0.309017101198944,-0.951056508751408);

		\draw[-,draw=red] (axis cs:1,0) -- (axis cs:1,0);
		\draw[-,draw=red] (axis cs:0.309016952187328,0.951056568356054) -- (axis cs:0.309016952187328,0.951056568356054);
		\draw[-,draw=red] (axis cs:-0.70901705649546,0.587785261505403) -- (axis cs:-0.70901705649546,0.587785261505403);
		\draw[-,draw=red] (axis cs:-0.709016996890813,-0.58778532111005) -- (axis cs:-0.709016996890813,-0.58778532111005);
		\draw[-,draw=red] (axis cs:0.309017101198944,-0.951056508751408) -- (axis cs:0.309017101198944,-0.951056508751408);
		\draw (axis cs:0.654513822846975,0.475520925003483) node[
			scale=0.45,
			fill=white,
			draw=white,
			line width=0.4pt,
			inner sep=3pt,
			text=black,
			rotate=307.5
		]{0.40};
		\draw (axis cs:0.1454882279220148,0.293893684737023) node[
			scale=0.45,
			fill=white,
			draw=white,
			line width=0.4pt,
			inner sep=3pt,
			text=black,
			rotate=342.9
		]{0.20};
		\draw (axis cs:-0.200021268149787,0.769414021435153) node[
			scale=0.45,
			fill=white,
			draw=white,
			line width=0.4pt,
			inner sep=3pt,
			text=black,
			rotate=17.1
		]{0.40};
		\draw (axis cs:-0.199989876539404,0.181649588137265) node[
			scale=0.45,
			fill=white,
			draw=white,
			line width=0.4pt,
			inner sep=3pt,
			text=black,
			rotate=52.5
		]{0.40};
		\draw (axis cs:0.309017026693093,5.8538300384825e-07) node[
			scale=0.45,
			fill=white,
			draw=white,
			line width=0.4pt,
			inner sep=3pt,
			text=black,
			rotate=270.0
		]{0.40};
		\draw (axis cs:-0.709017026693278,2.76257541953484e-06) node[
			scale=0.45,
			fill=white,
			draw=white,
			line width=0.4pt,
			inner sep=3pt,
			text=black,
			rotate=270.0
		]{0.20};
		\draw (axis cs:-0.200010123462488,-0.181621659109822) node[
			scale=0.45,
			fill=white,
			draw=white,
			line width=0.4pt,
			inner sep=3pt,
			text=black,
			rotate=307.5
		]{0.20};
		\draw (axis cs:1,0) node (499) [
			circle,
			scale=0.45,
			text=black,
			rotate=0.0,
			draw=black,
			fill=white,
		]{\tokens{499}};
		\draw (axis cs:0.309016952187328,0.951056568356054) node (699) [
			circle,
			scale=0.45,
			text=black,
			rotate=0.0,
			draw=black,
			fill=white,
		]{\tokens{699}};
		\draw (axis cs:-0.70901705649546,0.587785261505403) node (799) [
			circle,
			scale=0.45,
			text=black,
			rotate=0.0,
			draw=black,
			fill=white,
		]{\tokens{799}};
		\draw (axis cs:-0.709016996890813,-0.58778532111005) node (599) [
			circle,
			scale=0.45,
			text=black,
			rotate=0.0,
			draw=black,
			fill=white,
		]{\tokens{599}};
		\draw (axis cs:0.309017101198944,-0.951056508751408) node (299) [
			circle,
			scale=0.45,
			text=black,
			rotate=0.0,
			draw=black,
			fill=white,
		]{\tokens{299}};
    \draw[draw=red, line width=2.75pt] (699) to[loop above, looseness=5, in=0, out=120] node[
			midway,
			scale=0.45,
			fill=white,
			draw=white,
			line width=0.4pt,
			inner sep=3pt,
			text=black,
			rotate=0.0
		] {1.0} (699);
    		\draw[draw=red, line width=2.75pt] (499) to[loop above, looseness=5, in=40, out=-60] node[
			midway,
			scale=0.45,
			fill=white,
			draw=white,
			line width=0.4pt,
			inner sep=3pt,
			text=black,
			rotate=0.0
		] {1.0} (499);
    		\draw[draw=red, line width=2.75pt] (799) to[loop above, looseness=5, in=-140, out=120] node[
			midway,
			scale=0.45,
			fill=white,
			draw=white,
			line width=0.4pt,
			inner sep=3pt,
			text=black,
			rotate=0.0
		] {1.0} (799);
        		\draw[draw=red, line width=2.75pt] (599) to[loop above, looseness=5, in=-140, out=120] node[
			midway,
			scale=0.45,
			fill=white,
			draw=white,
			line width=0.4pt,
			inner sep=3pt,
			text=black,
			rotate=0.0
		] {1.0} (599);
        		\draw[draw=red, line width=2.75pt] (299) to[loop above, looseness=5, in=-40, out=60] node[
			midway,
			scale=0.45,
			fill=white,
			draw=white,
			line width=0.4pt,
			inner sep=3pt,
			text=black,
			rotate=0.0
		] {1.0} (299);
	\end{groupplot}

\end{tikzpicture}}
		\caption*{\centering\small \textbf{Epistemic}\\ Non-Contradiction Graph}
	\end{subfigure}
	\hfill
	\begin{subfigure}{.32\textwidth}
		\centering
		\small
		\resizebox{\textwidth}{!}{
\begin{tikzpicture}

	\definecolor{darkgray176}{RGB}{176,176,176}
	\definecolor{salmon255102102}{RGB}{255,102,102}

	\begin{groupplot}[group style={group size=3 by 3}]
		\nextgroupplot[
		height=1.4\textwidth,
		width=1.6\textwidth,
		every axis plot/.append style={line width=2pt},
		every mark/.append style={solid}, very thick,
		hide x axis,
		hide y axis,
    font=\LARGE,
		scaled x ticks=manual:{}{\pgfmathparse{#1}},
		scaled y ticks=manual:{}{\pgfmathparse{#1}},
		tick align=outside,
		x grid style={darkgray176},
		xmajorticks=false,
		xmin=-1.1998963847427483, xmax=1.38994679093202,
		xtick style={color=black},
		xticklabels={},
		y grid style={darkgray176},
		ymajorticks=false,
		ymin=-1.35553366454046, ymax=1.75063937800048,
		ytick style={color=black},
		yticklabels={}
		]
		\path [draw=red, line width=2.75pt]
		(axis cs:0.9,0)
		--(axis cs:0.9,0);

		\path [draw=salmon255102102, line width=3.2pt]
		(axis cs:0.9,0)
		--(axis cs:0.309016952187328,0.951056568356054);

		\path [draw=red, line width=2.75pt]
		(axis cs:0.9,0)
		--(axis cs:-0.80901705649546,0.587785261505403);

		\path [draw=salmon255102102, line width=3.2pt]
		(axis cs:0.9,0)
		--(axis cs:-0.709016996890813,-0.48778532111005);

		\path [draw=red, line width=2.75pt]
		(axis cs:0.9,0)
		--(axis cs:0.309017101198944,-0.951056508751408);

		\path [draw=red, line width=2.75pt]
		(axis cs:0.309016952187328,0.951056568356054)
		--(axis cs:0.309016952187328,0.951056568356054);

		\path [draw=salmon255102102, line width=3.2pt]
		(axis cs:0.309016952187328,0.951056568356054)
		--(axis cs:-0.80901705649546,0.587785261505403);

		\path [draw=salmon255102102, line width=3.2pt]
		(axis cs:0.309016952187328,0.951056568356054)
		--(axis cs:-0.709016996890813,-0.48778532111005);

		\path [draw=red, line width=2.75pt]
		(axis cs:0.309016952187328,0.951056568356054)
		--(axis cs:0.309017101198944,-0.951056508751408);

		\path [draw=red, line width=2.75pt]
		(axis cs:-0.80901705649546,0.587785261505403)
		--(axis cs:-0.80901705649546,0.587785261505403);

		\path [draw=salmon255102102, line width=3.2pt]
		(axis cs:-0.80901705649546,0.587785261505403)
		--(axis cs:-0.709016996890813,-0.48778532111005);

		\path [draw=red, line width=2.75pt]
		(axis cs:-0.80901705649546,0.587785261505403)
		--(axis cs:0.309017101198944,-0.951056508751408);

		\path [draw=red, line width=2.75pt]
		(axis cs:-0.709016996890813,-0.48778532111005)
		--(axis cs:-0.709016996890813,-0.48778532111005);

		\path [draw=red, line width=2.75pt]
		(axis cs:-0.709016996890813,-0.48778532111005)
		--(axis cs:0.309017101198944,-0.951056508751408);

		\path [draw=red, line width=2.75pt]
		(axis cs:0.309017101198944,-0.951056508751408)
		--(axis cs:0.309017101198944,-0.951056508751408);

		\draw[-,draw=red] (axis cs:0.9,0) -- (axis cs:0.9,0);
		\draw[-,draw=red] (axis cs:0.309016952187328,0.951056568356054) -- (axis cs:0.309016952187328,0.951056568356054);
		\draw[-,draw=red] (axis cs:-0.80901705649546,0.587785261505403) -- (axis cs:-0.80901705649546,0.587785261505403);
		\draw[-,draw=red] (axis cs:-0.709016996890813,-0.48778532111005) -- (axis cs:-0.709016996890813,-0.48778532111005);
		\draw[-,draw=red] (axis cs:0.309017101198944,-0.951056508751408) -- (axis cs:0.309017101198944,-0.951056508751408);
		\draw (axis cs:0.604513822846976,0.475520925003483) node[
			scale=0.45,
			fill=white,
			draw=white,
			line width=0.4pt,
			inner sep=3pt,
			text=black,
			rotate=307.5
		]{0.80};
		\draw (axis cs:0.0454882279220164,0.293893684737023) node[
			scale=0.45,
			fill=white,
			draw=white,
			line width=0.4pt,
			inner sep=3pt,
			text=black,
			rotate=342.9
		]{1.00};
		\draw (axis cs:0.0954882577244458,-0.243893714539454) node[
			scale=0.45,
			fill=white,
			draw=white,
			line width=0.4pt,
			inner sep=3pt,
			text=black,
			rotate=17.1
		]{0.80};
		\draw (axis cs:0.604513897351631,-0.475520895201621) node[
			scale=0.45,
			fill=white,
			draw=white,
			line width=0.4pt,
			inner sep=3pt,
			text=black,
			rotate=52.5
		]{1.00};
		\draw (axis cs:-0.250021268149787,0.769414021435153) node[
			scale=0.45,
			fill=white,
			draw=white,
			line width=0.4pt,
			inner sep=3pt,
			text=black,
			rotate=17.1
		]{0.80};
		\draw (axis cs:-0.199989876539404,0.231649588137265) node[
			scale=0.45,
			fill=white,
			draw=white,
			line width=0.4pt,
			inner sep=3pt,
			text=black,
			rotate=52.5
		]{0.80};
		\draw (axis cs:-0.759017026693277,0.05) node[
			scale=0.45,
			fill=white,
			draw=white,
			line width=0.4pt,
			inner sep=3pt,
			text=black,
			rotate=270.0
		]{0.80};
		\draw (axis cs:-0.250010123462488,-0.181621659109822) node[
			scale=0.45,
			fill=white,
			draw=white,
			line width=0.4pt,
			inner sep=3pt,
			text=black,
			rotate=307.5
		]{1.00};
		\draw (axis cs:-0.199978731848516,-0.719427808424042) node[
			scale=0.45,
			fill=white,
			draw=white,
			line width=0.4pt,
			inner sep=3pt,
			text=black,
			rotate=342.9
		]{1.00};
		\draw (axis cs:0.9,0) node (an) [
			circle,
			scale=0.45,
			text=black,
			rotate=0.0,
			draw=black,
			fill=white,
		]{\tokens{ an}};
		\draw (axis cs:0.309016952187328,0.951056568356054) node (the) [
			circle,
			scale=0.45,
			text=black,
			rotate=0.0,
			draw=black,
			fill=white,
		]{\tokens{ the}};
		\draw (axis cs:-0.80901705649546,0.587785261505403) node (a) [
			circle,
			scale=0.45,
			text=black,
			rotate=0.0,
			draw=black,
			fill=white,
		]{\tokens{ a}};
		\draw (axis cs:-0.709016996890813,-0.48778532111005) node (ef) [
			circle,
			scale=0.45,
			text=black,
			rotate=0.0,
			draw=black,
			fill=white,
		]{\tokens{ EF}};
		\draw (axis cs:0.309017101198944,-0.951056508751408) node (empty) [
			circle,
			scale=0.45,
			text=black,
			rotate=0.0,
			draw=black,
			fill=white,
		]{\tokens{ }};    
		\draw[draw=red, line width=2.75pt] (an) to[loop above, looseness=5, in=40, out=290] node[
			midway,
			scale=0.45,
			fill=white,
			draw=white,
			line width=0.4pt,
			inner sep=3pt,
			text=black,
		] {1.0} (an);
    \draw[draw=red, line width=2.75pt] (the) to[loop above, looseness=5, in=130, out=20] node[
			midway,
			scale=0.45,
			fill=white,
			draw=white,
			line width=0.4pt,
			inner sep=3pt,
			text=black,
		] {1.0} (the);
		\draw[draw=red, line width=2.75pt] (a) to[loop above, looseness=5, in=-140, out=120] node[
			midway,
			scale=0.45,
			fill=white,
			draw=white,
			line width=0.4pt,
			inner sep=3pt,
			text=black,
		] {1.0} (a);
		\draw[draw=red, line width=2.75pt] (ef) to[loop above, looseness=5, in=-120, out=140] node[
			midway,
			scale=0.45,
			fill=white,
			draw=white,
			line width=0.4pt,
			inner sep=3pt,
			text=black,
		] {1.0} (ef);
		\draw[draw=red, line width=2.75pt] (empty) to[loop above, looseness=5, in=-50, out=210] node[
			midway,
			scale=0.45,
			fill=white,
			draw=white,
			line width=0.4pt,
			inner sep=3pt,
			text=black,
		] {1.0} (empty);
	\end{groupplot}

\end{tikzpicture}}
		\caption*{\centering\small \textbf{Aleatoric}\\ Non-Contradiction Graph}
	\end{subfigure}
	\hfill
	\begin{subfigure}{.32\textwidth}
		\centering
		\small
		\resizebox{\textwidth}{!}{
\begin{tikzpicture}

	\definecolor{darkgray176}{RGB}{176,176,176}
	\definecolor{lavender204204255}{RGB}{204,204,255}

	\begin{groupplot}[group style={group size=3 by 3}]
		\nextgroupplot[
		height=1.4\textwidth,
		width=1.6\textwidth,
		every axis plot/.append style={line width=2pt},
		every mark/.append style={solid}, very thick,
		hide x axis,
		hide y axis,
    font=\LARGE,
		scaled x ticks=manual:{}{\pgfmathparse{#1}},
		scaled y ticks=manual:{}{\pgfmathparse{#1}},
		tick align=outside,
		x grid style={darkgray176},
		xmajorticks=false,
		xmin=-1.198963847427483, xmax=1.38994679093202,
		xtick style={color=black},
		xticklabels={},
		y grid style={darkgray176},
		ymajorticks=false,
		ymin=-2.15553366454046, ymax=2.95063937800048,
		ytick style={color=black},
		yticklabels={}
		]
		\path [draw=red, line width=2.75pt]
		(axis cs:1,0)
		--(axis cs:1,0);

		\path [draw=red, line width=2.75pt]
		(axis cs:1,0)
		--(axis cs:0.309016952187328,0.951056568356054);

		\path [draw=red, line width=2.75pt]
		(axis cs:1,0)
		--(axis cs:-0.60901705649546,0.657785261505403);

		\path [draw=red, line width=2.75pt]
		(axis cs:1,0)
		--(axis cs:-0.809016996890813,-0.58778532111005);

		\path [draw=red, line width=2.75pt]
		(axis cs:1,0)
		--(axis cs:0.309017101198944,-0.951056508751408);

		\path [draw=red, line width=2.75pt]
		(axis cs:0.309016952187328,0.951056568356054)
		--(axis cs:0.309016952187328,0.951056568356054);

		\path [draw=lavender204204255, line width=1.6pt]
		(axis cs:0.309016952187328,0.951056568356054)
		--(axis cs:-0.60901705649546,0.657785261505403);

		\path [draw=lavender204204255, line width=1.6pt]
		(axis cs:0.309016952187328,0.951056568356054)
		--(axis cs:-0.809016996890813,-0.58778532111005);

		\path [draw=lavender204204255, line width=1.6pt]
		(axis cs:0.309016952187328,0.951056568356054)
		--(axis cs:0.309017101198944,-0.951056508751408);

		\path [draw=red, line width=2.75pt]
		(axis cs:-0.60901705649546,0.657785261505403)
		--(axis cs:-0.60901705649546,0.657785261505403);

		\path [draw=red, line width=2.75pt]
		(axis cs:-0.809016996890813,-0.58778532111005)
		--(axis cs:-0.809016996890813,-0.58778532111005);

		\path [draw=red, line width=2.75pt]
		(axis cs:0.309017101198944,-0.951056508751408)
		--(axis cs:0.309017101198944,-0.951056508751408);

		\draw[-,draw=red] (axis cs:1,0) -- (axis cs:1,0);
		\draw[-,draw=red] (axis cs:0.309016952187328,0.951056568356054) -- (axis cs:0.309016952187328,0.951056568356054);
		\draw[-,draw=red] (axis cs:-0.60901705649546,0.657785261505403) -- (axis cs:-0.60901705649546,0.657785261505403);
		\draw[-,draw=red] (axis cs:-0.809016996890813,-0.58778532111005) -- (axis cs:-0.809016996890813,-0.58778532111005);
		\draw[-,draw=red] (axis cs:0.309017101198944,-0.951056508751408) -- (axis cs:0.309017101198944,-0.951056508751408);
		\draw (axis cs:0.654513822846977,0.475520925003483) node[
			scale=0.45,
			fill=white,
			draw=white,
			line width=0.4pt,
			inner sep=3pt,
			text=black,
			rotate=0.0,
		]{1.00};
		\draw (axis cs:0.19549147175227,0.3288926307527015) node[
      rectangle,
			scale=0.45,
			fill=white,
			draw=white,
			line width=0.4pt,
			inner sep=3pt,
			text=black,
			rotate=0.0,
		]{1.00};
		\draw (axis cs:0.0954882577244458,-0.293893714539454) node[
      rectangle,
			scale=0.45,
			fill=white,
			draw=white,
			line width=0.4pt,
			inner sep=3pt,
			text=black,
			rotate=0.0,
		]{1.00};
		\draw (axis cs:0.654513897351631,-0.475520895201621) node[
      rectangle,
			scale=0.45,
			fill=white,
			draw=white,
			line width=0.4pt,
			inner sep=3pt,
			text=black,
			rotate=0.0,
		]{1.00};
		\draw (axis cs:0.309016952187328,1.1412678760668) node[
			scale=0.45,
			fill=white,
			draw=white,
			line width=0.4pt,
			inner sep=3pt,
			text=black,
			rotate=0.0,
		]{1.00};
		\draw (axis cs:-0.150000052154066,0.8044209149307285) node[
      rectangle,
			scale=0.45,
			fill=white,
			draw=white,
			line width=0.4pt,
			inner sep=3pt,
			text=black,
			rotate=0.0,
		]{0.40};
		\draw (axis cs:-0.249989876539405,0.181649588137265) node[
			scale=0.45,
			fill=white,
			draw=white,
			line width=0.4pt,
			inner sep=3pt,
			text=black,
			rotate=0.0,
		]{0.40};
		\draw (axis cs:0.309017026693093,5.8538300384825e-07) node[
			scale=0.45,
			fill=white,
			draw=white,
			line width=0.4pt,
			inner sep=3pt,
			text=black,
			rotate=0.0,
		]{0.40};
		\draw (axis cs:1,0) node (empty) [
			circle,
			scale=0.45,
			text=black,
			rotate=0.0,
			draw=black,
			fill=white,
		]{\tokens{ }};
		\draw[draw=red, line width=2.75pt] (empty) to[loop above, looseness=5, in=0, out=120] node[
			midway,
			scale=0.45,
			fill=white,
			draw=white,
			line width=0.4pt,
			inner sep=3pt,
			text=black,
			rotate=0.0,
		] {1.0} (empty);
		\draw (axis cs:0.309016952187328,1.251056568356054) node (september) [
			circle,
			scale=0.45,
			text=black,
			rotate=0.0,
			draw=black,
			fill=white,
		]{\tokens{ September}};
		\draw[draw=red, line width=2.75pt] (september) to[loop above, looseness=3, in=0, out=120] node[
			midway,
			scale=0.45,
			fill=white,
			draw=white,
			line width=0.4pt,
			inner sep=3pt,
			text=black,
			rotate=0.0,
		] {1.0} (september);
		\draw (axis cs:-0.60901705649546,0.657785261505403) node (july) [
			circle,
			scale=0.45,
			text=black,
			rotate=0.0,
			draw=black,
			fill=white,
		]{\tokens{ July}};
		\draw[draw=red, line width=2.75pt] (july) to[loop above, looseness=5, in=-160, out=120] node[
			midway,
			scale=0.45,
			fill=white,
			draw=white,
			line width=0.4pt,
			inner sep=3pt,
			text=black,
			rotate=0.0,
		] {1.0} (july);
		\draw (axis cs:-0.809016996890813,-0.78778532111005) node (february) [
			circle,
			scale=0.45,
			text=black,
			rotate=0.0,
			draw=black,
			fill=white,
		]{\tokens{ February}};
		\draw[draw=red, line width=2.75pt] (february) to[loop above, looseness=3, in=-120, out=-20] node[
			midway,
			scale=0.45,
			fill=white,
			draw=white,
			line width=0.4pt,
			inner sep=3pt,
			text=black,
			rotate=0.0,
		] {1.0} (february);
		\draw (axis cs:0.309017101198944,-1.351056508751408) node (june) [
			circle,
			scale=0.45,
			text=black,
			rotate=0.0,
			draw=black,
			fill=white,
		]{\tokens{ June}};
		\draw[draw=red, line width=2.75pt] (june) to[loop above, looseness=5, in=-140, out=120] node[
			midway,
			scale=0.45,
			fill=white,
			draw=white,
			line width=0.4pt,
			inner sep=3pt,
			text=black,
		] {1.0} (june);
	\end{groupplot}

\end{tikzpicture}}
		\caption*{\centering\small \textbf{Low uncertainty}\\ Non-Contradiction Graph}
	\end{subfigure}
	\\[1.6em]
	\begin{subfigure}{.32\textwidth}
		\centering
		\small
		\resizebox{\textwidth}{!}{
\begin{tikzpicture}[every node/.append style={font=\large}]

	\definecolor{darkgray176}{RGB}{176,176,176}
	\definecolor{darkorange25512714}{RGB}{255,127,14}
	\definecolor{forestgreen4416044}{RGB}{44,160,44}
	\definecolor{steelblue31119180}{RGB}{31,119,180}

	\begin{groupplot}[group style={group size=3 by 3}]
		\nextgroupplot[
			height=1.4\textwidth,
			width=1.6\textwidth,
			every axis plot/.append style={line width=2pt},
			every mark/.append style={solid}, very thick,
			tick align=outside,
			tick pos=left,
			xticklabel style={font=\large, rotate=30, anchor=east, align=center},
			yticklabel style={font=\large},
			x grid style={darkgray176},
			xmin=-0.54, xmax=2.54,
			xtick style={color=black},
			xtick={0,1,2},
			xticklabels={{\color{steelblue31119180}Conditional},{\color{darkorange25512714}Semantic},{\color{forestgreen4416044}KTC=\\[-4pt]\color{forestgreen4416044}$\text{Conditional} \times \text{Heat}$}},
			y grid style={darkgray176},
			ymin=0, ymax=1.05,
			ytick style={color=black}
		]
		\draw[draw=none,fill=steelblue31119180] (axis cs:-0.4,0) rectangle (axis cs:0.4,0.888188942137307);
		\draw[draw=none,fill=darkorange25512714] (axis cs:0.6,0) rectangle (axis cs:1.4,0.927490939445527);
		\draw[draw=none,fill=forestgreen4416044] (axis cs:1.6,0) rectangle (axis cs:2.4,0.815628537215171);
		\draw (axis cs:0,0.888188942137307) ++(0pt,2pt) node[
			anchor=south,
			text=black,
			rotate=0.0,
			font=\small
		]{0.89};
		\draw (axis cs:1,0.927490939445527) ++(0pt,2pt) node[
			anchor=south,
			text=black,
			rotate=0.0,
			font=\small
		]{0.93};
		\draw (axis cs:2,0.815628537215171) ++(0pt,2pt) node[
			anchor=south,
			text=black,
			rotate=0.0,
			font=\small
		]{0.82};
	\end{groupplot}

\end{tikzpicture}}
		\caption*{\centering\small \textbf{Epistemic}\\ Entropy at each step}
	\end{subfigure}
	\hfill
	\begin{subfigure}{.32\textwidth}
		\centering
		\small
		\resizebox{\textwidth}{!}{
\begin{tikzpicture}[every node/.append style={font=\large}]

	\definecolor{darkgray176}{RGB}{176,176,176}
	\definecolor{darkorange25512714}{RGB}{255,127,14}
	\definecolor{forestgreen4416044}{RGB}{44,160,44}
	\definecolor{steelblue31119180}{RGB}{31,119,180}

	\begin{groupplot}[group style={group size=3 by 3}]
		\nextgroupplot[
			height=1.4\textwidth,
			width=1.6\textwidth,
			every axis plot/.append style={line width=2pt},
			every mark/.append style={solid}, very thick,
			tick align=outside,
			tick pos=left,
			xticklabel style={font=\large, rotate=30, anchor=east, align=center},
			yticklabel style={font=\large},
			x grid style={darkgray176},
			xmin=-0.54, xmax=2.54,
			xtick style={color=black},
			xtick={0,1,2},
			xticklabels={{\color{steelblue31119180}Conditional},{\color{darkorange25512714}Semantic},{\color{forestgreen4416044}KTC=\\[-4pt]\color{forestgreen4416044}$\text{Conditional} \times \text{Heat}$}},
			y grid style={darkgray176},
			ymin=0, ymax=1.05,
			ytick style={color=black}
		]
		\draw[draw=none,fill=steelblue31119180] (axis cs:-0.4,0) rectangle (axis cs:0.4,0.526853134820861);
		\draw[draw=none,fill=darkorange25512714] (axis cs:0.6,0) rectangle (axis cs:1.4,0.289546869688429);
		\draw[draw=none,fill=forestgreen4416044] (axis cs:1.6,0) rectangle (axis cs:2.4,0.182956710001256);
		\draw (axis cs:0,0.526853134820861) ++(0pt,2pt) node[
			anchor=south,
			text=black,
			rotate=0.0,
			font=\small
		]{0.53};
		\draw (axis cs:1,0.289546869688429) ++(0pt,2pt) node[
			anchor=south,
			text=black,
			rotate=0.0,
			font=\small
		]{0.29};
		\draw (axis cs:2,0.182956710001256) ++(0pt,2pt) node[
			anchor=south,
			text=black,
			rotate=0.0,
			font=\small
		]{0.18};
	\end{groupplot}

\end{tikzpicture}}
		\caption*{\centering\small \textbf{Aleatoric}\\ Entropy at each step}
	\end{subfigure}
	\hfill
	\begin{subfigure}{.32\textwidth}
		\centering
		\small
		\resizebox{\textwidth}{!}{
\begin{tikzpicture}[every node/.append style={font=\large}]

	\definecolor{darkgray176}{RGB}{176,176,176}
	\definecolor{darkorange25512714}{RGB}{255,127,14}
	\definecolor{forestgreen4416044}{RGB}{44,160,44}
	\definecolor{steelblue31119180}{RGB}{31,119,180}

	\begin{groupplot}[group style={group size=3 by 3}]
		\nextgroupplot[
			height=1.4\textwidth,
			width=1.6\textwidth,
			every axis plot/.append style={line width=2pt},
			every mark/.append style={solid}, very thick,
			tick align=outside,
			tick pos=left,
			xticklabel style={font=\large, rotate=30, anchor=east, align=center},
			yticklabel style={font=\large},
			x grid style={darkgray176},
			xmin=-0.54, xmax=2.54,
			xtick style={color=black},
			xtick={0,1,2},
			xticklabels={{\color{steelblue31119180}Conditional},{\color{darkorange25512714}Semantic},{\color{forestgreen4416044}KTC=\\[-4pt]\color{forestgreen4416044}$\text{Conditional} \times \text{Heat}$}},
			y grid style={darkgray176},
			ymin=0, ymax=1.05,
			ytick style={color=black}
		]
		\draw[draw=none,fill=steelblue31119180] (axis cs:-0.4,0) rectangle (axis cs:0.4,0.0164084196986163);
		\draw[draw=none,fill=darkorange25512714] (axis cs:0.6,0) rectangle (axis cs:1.4,0.731893131667053);
		\draw[draw=none,fill=forestgreen4416044] (axis cs:1.6,0) rectangle (axis cs:2.4,0.0119987445342124);
		\draw (axis cs:0,0.0164084196986163) ++(0pt,2pt) node[
			anchor=south,
			text=black,
			rotate=0.0,
			font=\small
		]{0.02};
		\draw (axis cs:1,0.731893131667053) ++(0pt,2pt) node[
			anchor=south,
			text=black,
			rotate=0.0,
			font=\small
		]{0.73};
		\draw (axis cs:2,0.0119987445342124) ++(0pt,2pt) node[
			anchor=south,
			text=black,
			rotate=0.0,
			font=\small
		]{0.01};
	\end{groupplot}

\end{tikzpicture}}
		\caption*{\centering\small \textbf{Low uncertainty}\\ Entropy at each step}
	\end{subfigure}
	\caption[Illustration of KTC computation for three tokens]{\textbf{Illustration of KTC computation for three tokens from the example in \Cref{fig:teaser}.} \textbf{Upper:} candidate tokens and their likelihoods. \textbf{Middle:} the non-contradiction graph. \textbf{Lower:} the entropy at different stages of the computation. \textbf{Left:} epistemic uncertainty, where the model hesitates between contradicting prices, resulting in a sparse non-contradiction graph and a high KTC score. \textbf{Center:} aleatoric uncertainty, where the model hesitates between non-contradicting formulations, yielding high conditional entropy but low semantic entropy and thus a low KTC score. \textbf{Right:} low uncertainty, with low conditional entropy; despite high semantic entropy, scaling leads to a low KTC score.}
    \label{fig:explaining_ktc}
\end{figure}

\subsection{[Step A] Building a Non-Contradiction Graph over Candidate Tokens} \label{sec:non-contradiction_graph}

First, we build a non-contradiction graph over the candidate tokens $\left\{ \Delta^\delta_i(x), i \in \intset{1}{\delta}\right\}$. For brevity, we simply refer to them as the $\Delta_i$'s. This graph is undirected and edge-weighted. Each node corresponds to an index $i \in \intset{1}{\delta}$, and the edge weights are scalars in $[0, 1]$ representing the non-contradiction between candidates $\Delta_i$ and $\Delta_j$. Previous methods rely on an NLI model to assess entailment between candidate tokens \citep{nikitin_kernel_2024, fadeeva_factchecking_2024, duan_shifting_2024}. However, this model induces a high computational cost, which can sometimes even surpass that of the LLM generation itself~\citep{dentan_much_2026}.

We introduce an alternative, extremely fast approach based on the most common neighboring tokens on Wikipedia. For a given LLM and a set of languages, we first tokenize a large part of the Wikipedia corpus. Then, for any candidate $\Delta_i$ and token $t \in \mathcal{V}$, we count how many times $t$ is a neighbor of $\Delta_i$, meaning that it appears right before or right after $\Delta_i$ in the corpus. Then, for $\nu \in \mathbb{N}^*$, we define $\mathcal{N}^\nu(\Delta_i) \subset \mathcal{V}$ as the $\nu$ most frequent neighbors of $\Delta_i$. We use the number of common neighbors between candidates, denoted $\left| \mathcal{N}^\nu(\Delta_i) \cap \mathcal{N}^\nu(\Delta_j) \right|$, to define the non-contradiction score as

\begin{equation} \label{eq:non_contradiction}
    w_{i,j}(\nu) = 1 - \min \left[
        \frac{\left|
            \mathcal{N}^\nu(\Delta_i) \cap \mathcal{N}^\nu(\Delta_j)
        \right|}{
            \min\left(
                \left| \mathcal{N}^\nu(\Delta_i) \right|,
                \left| \mathcal{N}^\nu(\Delta_j) \right|
            \right)}
        \ , \ 1 - \mathds{1}(\Delta_i \preceq \Delta_j \text{ or } \Delta_j \preceq \Delta_i) 
    \right] ,
\end{equation}

where $\Delta_i \preceq \Delta_j$ denotes that $\Delta_i$ is a prefix of $\Delta_j$. Indeed, we consider that such tokens should have a non-contradiction score of $1$ because the LLM can generate the missing suffix at the next NTP step.

\paragraph{Motivation and Validation} 

We observe that two contradicting tokens such as \tokens{699} and \tokens{499} can replace each other in a text, so they will have similar neighboring sets, and a low non-contradiction score. Conversely, two tokens that are unrelated such as \tokens{ } and \tokens{September}, or two tokens that entail each other such as \tokens{an} and \tokens{the}, will have different neighboring sets, and a non-contradiction score close to $1$. The similarity of the neighboring sets indicates contradiction between tokens.

A potential limitation of this heuristic is that semantically related concepts, such as synonyms, hypernyms, or entities from the same category, may also have overlapping Wikipedia neighborhoods, resulting in an inappropriately low non-contradiction score. To test this hypothesis, we computed $w_{i,j}$ on 100 pairs of synonyms, of hypernyms, and of semantically close but non-hypernym word pairs. The results, reported in \Cref{app:wikipedia_heuristic}, show that our heuristic assigns substantially higher non-contradiction scores to synonyms and hypernyms than to other semantically related concepts. It therefore captures non-contradiction rather than semantic similarity, as it was designed for.

\subsection{[Steps B and D] Building the Semantic Kernel and Computing Von Neumann Entropy} \label{sec:steps_b_d}

Let $\mathcal{G}$ be the non-contradiction graph defined above, with nodes $\intset{1}{\delta}$ and adjacency matrix $\mA = [w_{i,j}]_{1 \le i,j \le \delta}$. Let $\mD \in \mathbb{R}^{\delta \times \delta}$ be the degree matrix of the graph. For any $\tau \in \mathbb{R}_+$, we compute the Laplacian $\mL$, the \textit{heat kernel} $\mkappa(\tau)$ of $\mathcal{G}$, and its normalized version $\mK(\tau)$:

\begin{equation}
    \mL = \mD - \mA \quad ; \quad \mkappa(\tau) = \exp(-\tau\cdot \mL) \quad ; \quad \mK(\tau) = \mkappa(\tau) / \Tr(\mkappa(\tau)).
\end{equation}

We refer to $\mK$ as the \textit{semantic kernel}, since it is built upon the semantics of candidate tokens via $\mathcal{G}$. Finally, we compute the Von Neumann entropy (VNE) of $\mK(\tau)$ (see \Cref{sec:background}) to get the semantic entropy. Let $\lambda_1(\mK), \dots, \lambda_\delta(\mK)$ be the eigenvalues of $\mK(\tau)$. We define Semantic Entropy as:\footnote{Note that this definition differs from that of~\citet{farquhar_detecting_2024}, and only aims to distinguish it from Predictive Entropy.}

\begin{equation}
    \text{Semantic Entropy} = \VNE(\mK(\tau)) = - \sum_{i = 1}^{\delta} \lambda_i(\mK) \cdot \log \lambda_i(\mK).
\end{equation}

\paragraph{Motivation}
The eigenvalues of the Laplacian are closely related to the connectivity of the graph. The Laplacian always has at least one eigenvalue equal to zero, associated with the eigenvector $(1, \dots, 1)$. The second-smallest eigenvalue of $\mL$, also known as the Fiedler value, is equal to the \textit{algebraic connectivity} of the graph, which reflects how well connected the graph is~\citep{fiedler_algebraic_1973}. Here, the connectivity of the non-contradiction graph reflects semantic uncertainty: the more connected the graph is, the more the model hesitates between similar tokens, and therefore the lower the uncertainty.

Furthermore, let $\lambda_1 \le \dots \le \lambda_\delta$ be the eigenvalues of $\mL$. Then, the eigenvalues of $\mkappa(\tau)$ are given by $(e^{-\tau \lambda_1}, \dots, e^{-\tau \lambda_\delta})$. Therefore, low semantic uncertainty leads to a high Fiedler value, meaning that all eigenvalues of $\mL$ are large except the first one, which is zero. Consequently, $(e^{-\tau \lambda_1}, \dots, e^{-\tau \lambda_\delta}) \simeq (1, 0, \dots, 0)$, leading to a Von Neumann entropy close to zero. Conversely, high semantic uncertainty leads to a low Fiedler value, meaning that $e^{-\tau \lambda_1} \simeq e^{-\tau \lambda_2} \simeq 1$. Therefore, there are at least two similarly large eigenvalues of $\mK$, leading to a high Von Neumann entropy.

\paragraph{Numerical Examples}

Consider a perfectly sparse non-contradiction graph with $\mA = \mI_\delta$. This corresponds to a very high semantic uncertainty, because all tokens contradict each other, with a non-contradiction score of zero. Therefore, we have $\mL = 0$, $\mkappa = \mI_\delta$, and $\lambda_1 = \dots = \lambda_\delta = 1 / \delta$. The Von Neumann entropy is then maximal, equal to $\log \delta$. Conversely, consider the scenario where $\mathcal{G}$ is a clique. This corresponds to very low semantic uncertainty, because tokens do not contradict each other at all. Then, the vectors $(1, -1, 0, \dots, 0)$, $(1, 0, -1, 0, \dots)$, $\dots$, $(1, 0, \dots, 0, -1)$ are eigenvectors of $\mL$ associated with eigenvalue $\delta$. Therefore, the eigenvalues of $\mkappa$ are $(1, e^{-\tau \delta}, \dots, e^{-\tau \delta})$, and the Von Neumann entropy is close to zero for any non-negligible value of $\tau$.

These two numerical examples illustrate that VNE is high in the case of high semantic uncertainty and low in the case of low semantic uncertainty, which is its intended behavior.

\subsection{[Step C] Incorporating the Conditional Distribution} \label{sec:scaling_predictive}

Let $\mkappa$ be the non-normalized \textit{semantic} kernel defined above. Let $\smash{q_i \coloneqq [f(x)]_{\Delta_{i}}}$ be the conditional likelihood of the $i$-th candidate token. The matrix $\mPi \coloneqq \text{diag}(q_i / \sum_k q_k)_{1 \le i \le \delta} \in \mathbb{R}^{\delta \times \delta}$ is a unit-trace PSD kernel representing the conditional likelihood of candidate tokens, without any semantic information. We compute a combined PSD kernel $\mmu$ that integrates both semantic and conditional kernels, and we use it to compute the final $\KTC$ value as the Von Neumann entropy:

\begin{equation} \label{eq:ktc_value}
    \mmu = \mPi^{1/2} \mkappa \mPi^{1/2} \quad ; \quad \mM = \mmu / \Tr(\mmu) \quad ; \quad \boxed{\KTC =  - \sum_{i = 1}^{\delta} \lambda_i(\mM) \cdot \log \lambda_i(\mM).}
\end{equation}

\paragraph{Motivation}

The heat kernel $\mkappa$ is a \textit{diffusion operator}. Indeed, for any signal $u_0 \in \mathbb{R}^\delta$ defined on the graph, $u(\tau) = \mkappa(\tau) u_0$ is a solution of the diffusion equation $\partial u / \partial \tau + \mL u = 0$. Therefore, with the eigendecomposition $\mkappa = \mV^\top \mLambda \mV$, the eigenvectors contained in $\mV$ are graph signals that are stable under the diffusion process. The matrix $\mPi$ is an operator that scales a node signal according to the probability of each node. Thus, $\mmu = (\mV \mPi^{1/2})^\top \mLambda (\mV \mPi^{1/2})$ scales these stable signals with respect to the conditional likelihood of each node, thereby incorporating $\mPi$ into the semantic kernel.

Another interpretation is to observe that the eigenvalues of $\mmu = \mPi^{1/2} \mkappa \mPi^{1/2}$ are the same as those of $\mkappa \mPi$. Yet, $\mkappa \mPi$ is the composed operator of scaling by the predictive distribution and diffusion. This composition acts as a gating mechanism. For instance, if either $\mPi$ or $\mkappa$ has rank 1, then the rank of $\mkappa \mPi$ is also equal to 1 and the eigenvalues of $\mkappa \mPi$ are all zero except one, leading to an entropy of zero. This means that either no conditional uncertainty ($\text{rank}(\mPi) = 1$, such as $q_1 = 1$ and $q_{i>1} = 0$, similar to column~3 in \Cref{fig:explaining_ktc}) or no semantic uncertainty ($\text{rank}(\mkappa) = 1$, such as $\mA = \mathbf{1}$, the all-one matrix, similar to column~2 in \Cref{fig:explaining_ktc}) leads to low \KTC. The final entropy can be high only if both conditional entropy and semantic entropy are high (similar to column~1 in \Cref{fig:explaining_ktc}).

\section{Related Works} \label{sec:related_works}

\paragraph{Factual, White-Box, and Instance-Specific UQ} We focus exclusively on factuality hallucinations in LLM outputs, as opposed to \textit{input-conflicting}, \textit{context-conflicting}, or \textit{faithfulness} hallucinations~\citep{zhang_sirens_2025, huang_survey_2025}. Moreover, we focus on \textit{white-box} approaches that require access to the model itself, with its logits and activations \citep{fadeeva_factchecking_2024, sriramanan_llm-check_2024, chen_llms_2024, kadavath_language_2022, duan_shifting_2024}, as opposed to \textit{black-box} approaches that only rely on the generated text \citep{kuhn_semantic_2023, nikitin_kernel_2024, lin_generating_2023}. Finally, our approach is instance-specific as it computes an uncertainty score for a given LLM generation, as opposed to population-specific approaches that assign a score to a group of generations answering the same prompt \citep{sriramanan_llm-check_2024, nikitin_kernel_2024, kuhn_semantic_2023}.

\paragraph{Comparison with Kernel-Based and Semantic Methods}
Our work builds upon kernel methods and semantic similarities for UQ in LLMs. \citet{nikitin_kernel_2024} introduce Kernel Language Entropy (KLE), a generation-level method constructing PSD kernels from semantic similarities between outputs, computing their Von Neumann entropy. However, KLE operates at the \textit{generation level} as do the works that built upon it, such as SGPU~\citep{hoche_semantic_2025} or SNNE \citep{nguyen_beyond_2025}. In contrast, KTC adapts this principle to the \textit{token level}, enabling claim-specific UQ. Furthermore, KTC replaces the NLI models with a lightweight Wikipedia neighbor frequency heuristic, and integrates the predictive uncertainty.

\paragraph{White-Box Claim-Level UQ Methods Used as Baselines}

In \Cref{sec:empirical_results}, we compare KTC to other white-box methods adapted for claim-level UQ. As done in~\citep{dentan_much_2026}, we include CCP \citep{fadeeva_factchecking_2024} and SAR~\citep{duan_shifting_2024}, both of which rely on Natural Language Inference (NLI) scores. We also evaluate likelihood-based methods, including Maximum Likelihood \citep{aichberger_rethinking_2025}, defined as the probability of the most likely token at each step, Token Likelihood \citep{guerreiro_looking_2023}, corresponding to the probability of the sampled token, and Token Entropy \citep{malinin_uncertainty_2021}. We did not include FOCUS~\citep{zhang_enhancing_2023}, despite its strong performance on several benchmarks~\citep{yeh_halluentity_2025, rykov_when_2025}, because it relies on attention scores, which are unavailable in modern implementations such as Flash-Attention~\citep{dao_flashattention_2022, dao_flashattention2_2024}, making it impractical for production.
\section{Empirical Results} \label{sec:empirical_results}

Our evaluation shows that KTC achieves competitive performance with state-of-the-art methods while being significantly faster, making it practical for real-time monitoring of LLM outputs in production. We also show that KTC outperforms all baselines in the high-precision regime, which is the most relevant setting for real-world applications. Finally, we show that KTC generalizes well across languages, achieving competitive performance in all four languages tested.

We evaluated our approach on two benchmarks (MUCH~\citep{dentan_much_2026} and Mu-SHROOM~\citep{vazquez_semeval2025_2025}) spanning four languages (English, French, German, and Spanish) and 16 LLMs, including models from the Llama 3~\citep{grattafiori_llama_2024}, Mistral~\citep{liu_ministral_2026}, Gemma 3~\citep{team_gemma_2025}, Falcon~\citep{almazrouei_falcon_2023}, Pythia~\citep{biderman_pythia_2023}, Bloom~\citep{workshop_bloom_2023}, CroissantLLM~\citep{faysse_croissantllm_2025}, and Qwen 2~\citep{yang_qwen2_2024} families. We compared KTC with comparable white-box claim-level UQ methods, including CCP, SAR, Maximum Likelihood, Token Likelihood, and Token Entropy (see \Cref{sec:related_works}). All CPU experiments were run on an Apple M4 Pro (48 GB RAM), while GPU experiments were performed on an HPC node equipped with NVIDIA A100 GPUs and AMD EPYC 7543 CPUs.

\paragraph{\textit{Main} and \textit{Validation} Datasets} Our experiments focus on the high-precision/low-recall regime. Indeed, this regime is typically the most relevant for real-world deployment, where high precision is required to keep the number of false positives manageable~\citep{valentin_costeffective_2024,wang_developing_2025}. Consequently, to reduce noise due to annotation quality, we restrict our evaluations to the manually annotated subsets of MUCH and Mu-SHROOM in English, French, German, and Spanish. This forms our \textit{main} evaluation dataset, with a total of 698 samples that are segmented into 9,359 claims using the claim segmenter provided in MUCH~\citep{dentan_much_2026}. Moreover, to assess the generalization of KTC at a larger scale and on samples it was not tuned for, we also use the 4,673 samples of MUCH that were not annotated by humans as our \textit{validation} set, comprising 19,884 claims.

\paragraph{Hyperparameters}

We evaluate all methods over predefined hyperparameter grids, summarized in \Cref{tab:hyperparameter_grid} in \Cref{app:hyperparameter_tuning}. Among the 30 configurations that we explored for KTC, we retained the best 8 according to ROC-AUC and PR-AUC. For CCP and SAR, we evaluated 18 and 9 parameter configurations that include three sizes of underlying NLI models (70M, 100M, and 350M) that we introduce to enable faster inference and assess whether this could improve their accuracy--efficiency trade-off relative to KTC. Hyperparameter generalization is discussed in \Cref{sec:hyperparameter_generalization}.

\subsection{Is Real-Time Monitoring of LLM Outputs Practical with KTC?}

\begin{figure}[t!]
	\hfill
	\begin{subfigure}[t]{.49\textwidth}
		\small
		\centering
		\resizebox{.97\textwidth}{!}{
\begin{tikzpicture}

\definecolor{crimson2143940}{RGB}{214,39,40}
\definecolor{darkgray176}{RGB}{176,176,176}
\definecolor{darkorange25512714}{RGB}{255,127,14}
\definecolor{forestgreen4416044}{RGB}{44,160,44}
\definecolor{lightgray204}{RGB}{204,204,204}
\definecolor{mediumpurple148103189}{RGB}{148,103,189}
\definecolor{steelblue31119180}{RGB}{31,119,180}

\begin{groupplot}[group style={group size=2 by 1}]
\nextgroupplot[
every mark/.append style={solid}, very thick,
legend cell align={left},
legend style={
  fill opacity=0.8,
  draw opacity=1,
  text opacity=1,
  at={(0.97,0.03)},
  anchor=south east,
  draw=lightgray204
},
log basis x={10},
tick align=outside,
tick pos=left,
x grid style={darkgray176},
xlabel={Proportion of LLM computation time},
xmajorgrids,
xmin=0.0182796248282595, xmax=53.6917203356727,
xminorgrids,
xmode=log,
xtick style={color=black},
y grid style={darkgray176},
ylabel={ROC-AUC},
ymajorgrids,
ymin=0.556053142808967, ymax=0.697754922905915,
yminorgrids,
ytick style={color=black}
]
\addplot [draw=steelblue31119180, fill=steelblue31119180, mark=*, only marks]
table{%
x  y
0.0566483084185681 0.679489601103061
0.0566483084185681 0.678557129506224
0.0566483084185681 0.678735103325267
0.0566483084185681 0.677921440389085
0.0566483084185681 0.67248240142977
0.0566483084185681 0.676735267158045
0.0566483084185681 0.674202944565326
0.0566483084185681 0.675097148651937
};
\addlegendentry{KTC (Ours)}
\addplot [draw=darkorange25512714, fill=darkorange25512714, mark=x, only marks, opacity=0.3]
table{%
x  y
0.881195908733281 0.680944603593439
0.461054287962234 0.666951400859943
0.398111723052714 0.564442218152313
1.02596380802518 0.688539927064517
0.492525570416995 0.661553798439044
0.403619197482297 0.562740958254332
1.14398111723053 0.685892921527089
0.544453186467349 0.643706399702948
0.407553107789142 0.562494132813374
1.15892997639654 0.682767602008327
0.549960660896932 0.673028042311908
0.43273013375295 0.570896960543207
1.34697088906373 0.691313932901508
0.618410700236035 0.67025825188034
0.447678992918961 0.568635589860784
1.62627852084972 0.689959621797413
0.720692368214005 0.649951956336771
0.486231313926042 0.568580161763941
};
\addlegendentry{CCP (gpu)}
\addplot [draw=darkorange25512714, fill=darkorange25512714, mark=*, only marks]
table{%
x  y
11.4870180959874 0.680944603593439
3.69000786782061 0.666951400859943
1.8882769472856 0.564442218152313
14.6420141620771 0.688539927064517
4.57120377655389 0.661553798439044
2.2029897718332 0.562740958254332
18.4893784421715 0.685892921527089
5.86939417781275 0.643706399702948
2.65145554681353 0.562494132813374
20.6294256490952 0.682767602008327
6.56963021243116 0.673028042311908
3.75295043273013 0.570896960543207
27.5059008654603 0.691313932901508
8.59952793076318 0.67025825188034
4.83870967741935 0.568635589860784
37.3485444531865 0.689959621797413
11.6443745082612 0.649951956336771
7.18332022029898 0.568580161763941
};
\addlegendentry{CCP (cpu)}
\addplot [draw=forestgreen4416044, fill=forestgreen4416044, mark=x, only marks, opacity=0.3]
table{%
x  y
0.110936270653029 0.67207027828253
0.0637293469708891 0.667258019285629
0.0550747442958301 0.621530497112896
0.131392604248623 0.676531253494153
0.0708103855232101 0.6640033675409
0.0574350904799371 0.616070680091388
0.162863886703383 0.673484382371354
0.0818253343823761 0.66161140897983
0.0590086546026751 0.615766274006136
};
\addlegendentry{SAR (gpu)}
\addplot [draw=forestgreen4416044, fill=forestgreen4416044, mark=*, only marks]
table{%
x  y
2.16365066876475 0.67207027828253
0.794649881982691 0.667258019285629
0.613690007867821 0.621530497112896
2.94256490952006 0.676531253494153
1.03068450039339 0.6640033675409
0.692368214004721 0.616070680091388
4.09913453973249 0.673484382371354
1.34539732494099 0.66161140897983
0.818253343823761 0.615766274006136
};
\addlegendentry{SAR (cpu)}
\addplot [draw=crimson2143940, fill=crimson2143940, mark=*, only marks]
table{%
x  y
0.0278520849724626 0.663477129482306
0.0278520849724626 0.669753419710279
0.0276947285601888 0.672986665566476
};
\addlegendentry{Entropy}
\addplot [draw=mediumpurple148103189, fill=mediumpurple148103189, mark=*, only marks]
table{%
x  y
0.0264358772619984 0.660751645652422
0.0262785208497246 0.669176895751529
};
\addlegendentry{Likelihood}
\end{groupplot}

\end{tikzpicture}}
		\caption{ROC-AUC}
	\end{subfigure}
	\hfill
	\begin{subfigure}[t]{.49\textwidth}
		\small
		\centering
		\resizebox{.97\textwidth}{!}{
\begin{tikzpicture}

\definecolor{crimson2143940}{RGB}{214,39,40}
\definecolor{darkgray176}{RGB}{176,176,176}
\definecolor{darkorange25512714}{RGB}{255,127,14}
\definecolor{forestgreen4416044}{RGB}{44,160,44}
\definecolor{lightgray204}{RGB}{204,204,204}
\definecolor{mediumpurple148103189}{RGB}{148,103,189}
\definecolor{steelblue31119180}{RGB}{31,119,180}

\begin{groupplot}[group style={group size=2 by 1}]
\nextgroupplot[
every mark/.append style={solid}, very thick,
legend cell align={left},
legend style={
  fill opacity=0.8,
  draw opacity=1,
  text opacity=1,
  at={(0.97,0.03)},
  anchor=south east,
  draw=lightgray204
},
log basis x={10},
tick align=outside,
tick pos=left,
x grid style={darkgray176},
xlabel={Proportion of LLM computation time},
xmajorgrids,
xmin=0.0182796248282595, xmax=53.6917203356727,
xminorgrids,
xmode=log,
xtick style={color=black},
y grid style={darkgray176},
ylabel={Partial PR-AUC (REC<20\%)},
ymajorgrids,
ymin=0.320043449847876, ymax=0.802738157614531,
yminorgrids,
ytick style={color=black}
]
\addplot [draw=steelblue31119180, fill=steelblue31119180, mark=*, only marks]
table{%
x  y
0.0566483084185681 0.748274821915862
0.0566483084185681 0.759038524307994
0.0566483084185681 0.754012486689679
0.0566483084185681 0.780797489079683
0.0566483084185681 0.770390540551735
0.0566483084185681 0.751525750358317
0.0566483084185681 0.779367139560926
0.0566483084185681 0.743443103401386
};
\addlegendentry{KTC (Ours)}
\addplot [draw=darkorange25512714, fill=darkorange25512714, mark=x, only marks, opacity=0.3]
table{%
x  y
0.881195908733281 0.592267495406347
0.461054287962234 0.631605785356822
0.398111723052714 0.344888998457484
1.02596380802518 0.621462624785015
0.492525570416995 0.655469423501253
0.403619197482297 0.341984118382724
1.14398111723053 0.646869154416611
0.544453186467349 0.676214314707284
0.407553107789142 0.346323015809486
1.15892997639654 0.606705814730861
0.549960660896932 0.646816349439082
0.43273013375295 0.349252415378434
1.34697088906373 0.633400420931743
0.618410700236035 0.66961315983536
0.447678992918961 0.347228813042021
1.62627852084972 0.656210566390608
0.720692368214005 0.685728499269664
0.486231313926042 0.351619452712697
};
\addlegendentry{CCP (gpu)}
\addplot [draw=darkorange25512714, fill=darkorange25512714, mark=*, only marks]
table{%
x  y
11.4870180959874 0.592267495406347
3.69000786782061 0.631605785356822
1.8882769472856 0.344888998457484
14.6420141620771 0.621462624785015
4.57120377655389 0.655469423501253
2.2029897718332 0.341984118382724
18.4893784421715 0.646869154416611
5.86939417781275 0.676214314707284
2.65145554681353 0.346323015809486
20.6294256490952 0.606705814730861
6.56963021243116 0.646816349439082
3.75295043273013 0.349252415378434
27.5059008654603 0.633400420931743
8.59952793076318 0.66961315983536
4.83870967741935 0.347228813042021
37.3485444531865 0.656210566390608
11.6443745082612 0.685728499269664
7.18332022029898 0.351619452712697
};
\addlegendentry{CCP (cpu)}
\addplot [draw=forestgreen4416044, fill=forestgreen4416044, mark=x, only marks, opacity=0.3]
table{%
x  y
0.110936270653029 0.639131678984142
0.0637293469708891 0.644949508067962
0.0550747442958301 0.441969801701472
0.131392604248623 0.65060503008018
0.0708103855232101 0.657834223646135
0.0574350904799371 0.428605201066027
0.162863886703383 0.645102627389704
0.0818253343823761 0.671530438495631
0.0590086546026751 0.423044271356856
};
\addlegendentry{SAR (gpu)}
\addplot [draw=forestgreen4416044, fill=forestgreen4416044, mark=*, only marks]
table{%
x  y
2.16365066876475 0.639131678984142
0.794649881982691 0.644949508067962
0.613690007867821 0.441969801701472
2.94256490952006 0.65060503008018
1.03068450039339 0.657834223646135
0.692368214004721 0.428605201066027
4.09913453973249 0.645102627389704
1.34539732494099 0.671530438495631
0.818253343823761 0.423044271356856
};
\addlegendentry{SAR (cpu)}
\addplot [draw=crimson2143940, fill=crimson2143940, mark=*, only marks]
table{%
x  y
0.0278520849724626 0.6711744389313
0.0278520849724626 0.703971132742388
0.0276947285601888 0.695618761748085
};
\addlegendentry{Entropy}
\addplot [draw=mediumpurple148103189, fill=mediumpurple148103189, mark=*, only marks]
table{%
x  y
0.0264358772619984 0.450104010931969
0.0262785208497246 0.685115211860505
};
\addlegendentry{Likelihood}
\end{groupplot}

\end{tikzpicture}}
		\caption{PR-AUC focusing on High Precision}
	\end{subfigure}
	\hfill
    \caption{\textbf{Accuracy-efficiency trade-off for white-box claim-level UQ methods.} Each point corresponds to a method evaluated under a different set of hyperparameters. The $x$-axis reports the UQ computation time as a fraction of the LLM generation cost. KTC achieves competitive accuracy at a fraction of the cost of NLI-based methods (CCP, SAR), while requiring no GPU.}
	\label{fig:compute_time}
\end{figure}
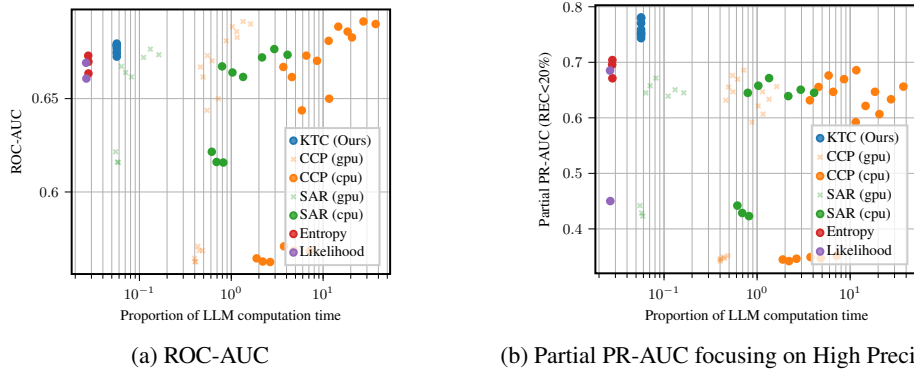

The speed of a UQ method is crucial for real-time monitoring applications. It should be quantified as a computational overhead relative to the LLM inference time~\citep{dentan_much_2026}. If a method requires as much time or more than the LLM inference, its use becomes prohibitive in practice, as it would significantly increase the overall latency of the system. In contrast, a method with a negligible overhead can be easily integrated into existing pipelines without causing significant delays. To this end, we evaluate the performance of KTC and the baselines as a function of the computational overhead relative to the LLM inference time (see \Cref{fig:compute_time}). The overhead is computed as the ratio of the time taken by the UQ to the inference runtime, averaged over all samples, models, and languages. In this figure, we report the area under the curve (AUC) for the ROC curve, and for the precision-recall (PR) curve focusing on the high-precision regime (recall $\leq 20\%$). For NLI-based methods (CCP and SAR), we report the overhead for both CPU and GPU implementations, as these methods can be significantly accelerated by GPU computation. For KTC and other baselines, we report the overhead for the CPU implementation, as they do not require GPU acceleration. We include multiple points per method, corresponding to different hyperparameter sets (see details in \Cref{app:hyperparameter_tuning}).

KTC achieves the best overall trade-off between performance and efficiency. In terms of ROC-AUC, the only baseline that surpasses KTC is CCP. However, KTC reaches its top performance with only a $5.6\%$ overhead and runs entirely on CPU. In contrast, the CCP configurations that achieve comparable performance incur a $46\%$ overhead on GPU (8.2$\times$ higher) and up to $369\%$ on CPU (65$\times$ higher). For PR-AUC, KTC not only outperforms all baselines but does so by a clear margin in the high-precision regime. It achieves a PR-AUC of $0.78$ at recall $\leq 20\%$, whereas CCP peaks at $0.69$, despite its much higher cost. Moreover, the introduction of the 70M NLI variant for CCP does not improve the trade-off between performance and computational efficiency. This shows that the higher performance of CCP requires a high-performing NLI model, which incurs a high computational cost. Other methods further illustrate the favorable trade-off of KTC: SAR incurs a higher overhead ($\sim 10\%$ of the LLM generation cost on GPU) while delivering lower performance, and entropy- and likelihood-based baselines, although cheaper ($\sim 2.7\%$ overhead compared to LLM generation), remain behind in both ROC-AUC and PR-AUC. Overall, KTC is the only method that combines strong performance with consistently low computational cost.

\subsection{Does KTC Generalize across Languages?}

\begin{figure}[h!]
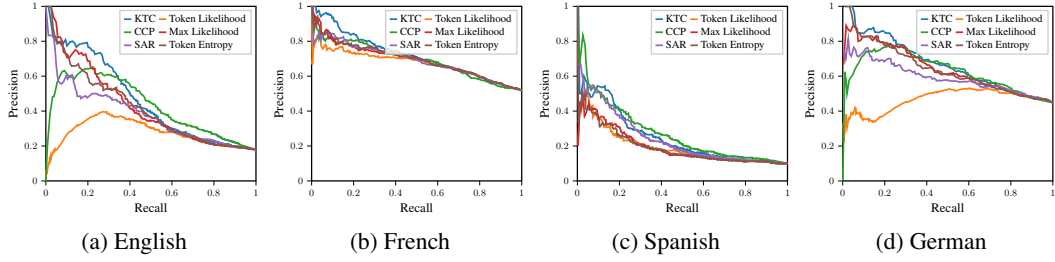

	\centering
	\hfill
	\begin{subfigure}{.245\textwidth}
		\centering
		\small
		\resizebox{\textwidth}{!}{\input{figures/results/pr_per_lang/pr_per_lang_0_0.tikz}}
		\caption{English}
	\end{subfigure}
	\hfill
	\begin{subfigure}{.245\textwidth}
		\centering
		\small
		\resizebox{\textwidth}{!}{\input{figures/results/pr_per_lang/pr_per_lang_1_0.tikz}}
		\caption{French}
	\end{subfigure}
	\hfill
	\begin{subfigure}{.245\textwidth}
		\centering
		\small
		\resizebox{\textwidth}{!}{\input{figures/results/pr_per_lang/pr_per_lang_0_1.tikz}}
		\caption{Spanish}
	\end{subfigure}
	\begin{subfigure}{.245\textwidth}
		\centering
		\small
		\resizebox{\textwidth}{!}{\input{figures/results/pr_per_lang/pr_per_lang_1_1.tikz}}
		\caption{German}
	\end{subfigure}
	\hfill

	\caption{\textbf{Precision-Recall curves for claim-level hallucination detection} across four languages, using product as token-to-claim aggregation method. Results are averaged over 16 LLMs across two benchmarks. KTC notably outperforms competing methods in the high-precision regime (low recall), while remaining competitive overall.}\label{fig:pr_per_lang}
\end{figure}

We compare the Precision-Recall curves of KTC and our baselines across four languages in \Cref{fig:pr_per_lang} (see ROC curves in \Cref{app:roc_curves}). To generate those figures, we evaluate all hyperparameter combinations and report results for the best hyperparameter set for each method and language. KTC consistently outperforms all baselines across languages in the low-recall, high-precision regime, which is typically the most relevant in production (see above). This trend holds in all languages and is particularly pronounced in English, French, and German, where KTC maintains a clear precision advantage over all baselines up to a recall of $0.4$. In Spanish, the gap is narrower in the high-precision regime, and CCP surpasses KTC beyond $20\%$ recall. We attribute this to a lower prevalence of hallucinations in Spanish, which makes the high-precision regime inherently more challenging.

Overall, KTC delivers competitive performance across all languages while consistently preserving its advantage in the high-precision regime. Notably, KTC relies on a Wikipedia-based non-contradiction graph that is language-agnostic. In contrast, top-performing baselines such as SAR and CCP depend on NLI models trained in English, which can limit their ability to generalize across languages.

\subsection{Does KTC Generalize to the Validation Set?} \label{sec:hyperparameter_generalization}

We evaluated KTC and our baselines with various hyperparameters (see above). In \Cref{fig:compute_time}, we retain the 8 top-performing KTC configurations according to ROC-AUC and PR-AUC. Most use $\tau\in\{0.2,0.3\}$ and $\nu\in\{3,4,5\}$, which we recommend as default values leading to similarly good performance. To assess hyperparameter generalization, we additionally evaluate these 8 configurations, without retuning, on our \textit{validation} set introduced above. We compare them against all baseline configurations. Our results are reported in \Cref{fig:compute_time_test_much} and \Cref{tab:compute_time_test_much_table}. KTC maintains its favorable accuracy-efficiency trade-off on this unseen set, supporting the generalization of the selected hyperparameters at larger scale.

\subsection{What is the Impact of Token-to-Claim Aggregation?}

\begin{figure}[h!]
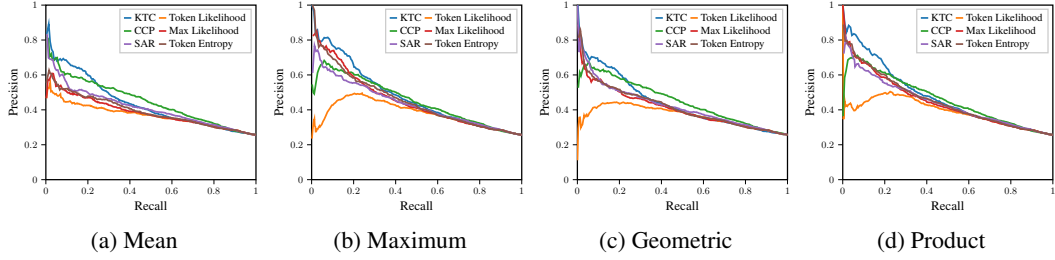

	\centering
	\hfill
	\begin{subfigure}{.245\textwidth}
		\centering
		\small
		\resizebox{\textwidth}{!}{\input{figures/results/pr_per_agg/pr_per_agg_0_0.tikz}}
		\caption{Mean}
	\end{subfigure}
	\hfill
	\begin{subfigure}{.245\textwidth}
		\centering
		\small
		\resizebox{\textwidth}{!}{\input{figures/results/pr_per_agg/pr_per_agg_1_0.tikz}}
		\caption{Maximum}
	\end{subfigure}
	\hfill
	\begin{subfigure}{.245\textwidth}
		\centering
		\small
		\resizebox{\textwidth}{!}{\input{figures/results/pr_per_agg/pr_per_agg_0_1.tikz}}
		\caption{Geometric}
	\end{subfigure}
	\begin{subfigure}{.245\textwidth}
		\centering
		\small
		\resizebox{\textwidth}{!}{\input{figures/results/pr_per_agg/pr_per_agg_1_1.tikz}}
		\caption{Product}
	\end{subfigure}
	\hfill
    \caption{\textbf{Precision-recall curves under four token-to-claim aggregation strategies} (Mean, Maximum, Geometric mean, Product), averaged across four languages and 16 LLMs. Each method is evaluated over its full grid search space, and we report its best-performing configuration.}\label{fig:pr_per_agg}
\end{figure}

KTC produces an uncertainty score at each token position within a claim. To obtain a single claim-level score, we aggregate these per-position scores using one of four strategies: mean, maximum, geometric mean, or product over the token-level KTC values. \Cref{fig:pr_per_agg} compares KTC against all baselines under each strategy, averaged over all four languages and 16 LLMs. Further results are provided in \Cref{fig:roc_per_agg} in \Cref{app:roc_curves}. Across every aggregation strategy, KTC consistently dominates all baselines in the high-precision regime and achieves competitive ROC-AUC, confirming that the method's advantage is not contingent on a particular aggregation choice. We use the product aggregation as the default throughout the paper, as it is the one giving the best performance for all methods, aligning with the findings of~\citet{fadeeva_factchecking_2024}.

\subsection{What is the Impact of the Semantic and Predictive Kernels?}

We propose two approaches for merging the \textit{predictive kernel} $\mPi$ and the \textit{semantic kernel} $\mkappa$:
\begin{equation}
    \mmu^\text{SCALE} = \mPi^{1/2} \mkappa \mPi^{1/2}
    \quad \quad \text{and} \quad \quad
    \mmu^\text{COMB}_\alpha = \alpha \mPi + (1-\alpha) \mkappa, \quad \alpha \in [0, 1].
\end{equation}

In both cases, we divide the matrix by its trace to obtain $\mM^\text{SCALE}$ and $\mM^\text{COMB}_\alpha$ before computing the VNE, as in \Cref{eq:ktc_value}. $\mM^\text{SCALE}$ is simply denoted $\mM$ in the rest of this paper because we use the SCALE method as the default for three reasons. First, it consistently outperforms COMB as evidenced in \Cref{tab:ablation_agg}. Second, it does not rely on any hyperparameter. Finally, it can be interpreted intuitively as a scaling of the kernel $\mkappa$ by the conditional probabilities of the candidate tokens. In contrast, the convex combination involved in COMB is less interpretable.

The ablation of the $\alpha$ parameter for the COMB strategy in \Cref{tab:ablation_agg} provides insight into the respective roles of the semantic and predictive kernels. When $\alpha = 0$, the resulting kernel reduces to the purely semantic kernel and performs significantly worse. Increasing $\alpha$ improves performance, indicating that the conditional kernel plays a more dominant role. The fact that SCALE outperforms all COMB values shows that the predictive kernel alone is still insufficient, highlighting the benefit of the SCALE method for combining both components.

\begin{table}[h!]
    \caption{\textbf{Precision-Recall AUC of aggregation methods} for varying neighborhood size $\nu$. SCALE consistently outperforms COMB across all configurations.}
    \label{tab:ablation_agg}
    \centering
    \begin{tabular}{ccccccc}
        \toprule
        \multirow{2}{*}{\textbf{$\nu$}} & \multirow{2}{*}{\textbf{SCALE}} & \multicolumn{5}{c}{\textbf{COMB}} \\
        \cmidrule(lr){3-7}
        & & $\alpha = 0$ & $\alpha = 0.25$ & $\alpha = 0.5$ & $\alpha = 0.75$ & $\alpha = 1$ \\
        \midrule
        3 & \textbf{0.471} & 0.402 & 0.398 & 0.410 & 0.432 & 0.452 \\
        4 & \textbf{0.476} & 0.404 & 0.400 & 0.412 & 0.433 & 0.452 \\
        5 & \textbf{0.472} & 0.395 & 0.395 & 0.409 & 0.431 & 0.452 \\
        6 & \textbf{0.467} & 0.391 & 0.392 & 0.407 & 0.431 & 0.452 \\
        8 & \textbf{0.457} & 0.379 & 0.383 & 0.401 & 0.428 & 0.452 \\
        \bottomrule
    \end{tabular}
\end{table}

\section{Limitations and Future Work} \label{sec:limitations}

Our evaluation is limited to two manually annotated benchmarks (698 samples in total), which ensures high quality but remains relatively small-scale. We also restrict our experiments to English, French, German, and Spanish due to reliance on the MUCH segmenter~\citep{dentan_much_2026}; extending to more languages would be a natural next step. Moreover, KTC does not dominate all baselines across all regimes, but instead provides a stronger performance--computation trade-off and consistently outperforms other methods in the high-precision setting, which is often the most relevant in practice.

Finally, KTC assumes access to token-level probabilities. Extending it to black-box settings with partial or approximate log-probabilities is an important direction for future work.

\section{Conclusion} \label{sec:conclusion}

We introduced KTC, a token-level uncertainty quantification method that combines predictive uncertainty with a contradiction signal derived from Wikipedia neighbor statistics. By replacing the NLI models used in prior work with a lightweight corpus-based heuristic, KTC removes the dependency on GPU acceleration and English-trained inference models that limits existing approaches.

Empirically, KTC ranks second to CCP on overall ROC-AUC but achieves the best accuracy--cost trade-off by a wide margin, and clearly dominates all baselines in the high-precision regime that matters most for production hallucination mitigation. These results hold across four languages and 16 LLMs. While KTC requires white-box access to token probabilities and our evaluation covers a focused set of languages and benchmarks, its combination of CPU-only efficiency and high-precision accuracy makes it, to our knowledge, the first claim-level UQ method practical for real-time monitoring of LLM outputs under production latency constraints.

\begin{ack}
This work received direct financial support from Crédit Agricole SA through the research chair "Trustworthy and Responsible AI" with École polytechnique. This work was granted access to the HPC resources of IDRIS under the allocation 2023-AD011014843R2 made by GENCI. We also thank Mohamed Dhouib and Mathis Le Bail for discussions on early versions of this paper. 

Regarding competing interests, the authors declare ongoing research collaborations with Groupe SNCF and paid work by one author for Mistral AI within the 36 months preceding submission. These relationships did not influence the work presented in this paper.
\end{ack}

\printbibliography[heading=bibintoc, title={Bibliography}]

\FloatBarrier
\appendix
\crefalias{section}{appendix}
\section{Choice of Kernel}

The choice of kernel is crucial to accurately capture semantic uncertainty. While we used a heat kernel in our method, we also experimented with other kernels, introduced below.

Let $\mathcal{G}$ be the non-contradiction graph defined above, with nodes $\intset{1}{\delta}$ and adjacency matrix $\mA = [w_{i,j}]_{1 \le i,j \le \delta}$. Let $\mD \in \mathbb{R}^{\delta \times \delta}$ be the degree matrix of the graph. Given that $\mA$ is symmetric, we also consider its eigendecomposition $\mA = \mU\mSigma\mU^\top$, where $\mSigma$ is a diagonal matrix containing the eigenvalues. We consider several approaches to build a PSD kernel upon $\mA$:

\begin{equation}
	\left\{
	\begin{aligned}
		\mkappa^\text{CLIP}       & = \mU \times \text{clip}(\mSigma, \min=0) \times \mU^\top \\
		\mkappa^\text{AUGM}       & = \mA - \min(0, \min(\mSigma)) \times \mI_\delta   \\
		\mkappa^\text{SQR}        & = \mA \times \mA                                         \\
		\mkappa^\text{LAPL}       & = \mD - \mA \\
		\mkappa^\text{HEAT}(\tau) & = \exp(-\tau\cdot [\mD - \mA]), \quad \tau \in ]0, 1]
	\end{aligned}
	\right.
\end{equation}

where $\text{clip}(\cdot)$ is an element-wise operations and $\min(\cdot)$ is a scalar denoting the minimum. For each of these kernels, we divide it by its trace, leading to the unit-trace PSD kernels $\mK^\text{CLIP}$, $\mK^\text{AUGM}$, $\mK^\text{SQR}$, $\mK^\text{LAPL}$, and $\mK^\text{HEAT}=\mK$. All of these kernels are built on the semantics of the candidate tokens via the non-contradiction graph, which is why we refer to these kernels as the \textit{semantic} kernels.

In practice, preliminary experiments with KTC showed that the other kernels are less effective than the heat kernel at capturing semantic uncertainty, in line with the findings of \citet{nikitin_kernel_2024}. Consequently, we exclude them from our experiments and focus solely on the heat kernel.
\section{Validation of the Wikipedia Heuristic}
\label{app:wikipedia_heuristic}

KTC requires a measure of whether two candidate tokens can be substituted without changing the factual content of the generated statement. We estimate this quantity from Wikipedia rather than using a neural semantic similarity model, which keeps the method lightweight and CPU-only. For each token, we construct a neighborhood from its occurrences in Wikipedia. The non-contradiction score between two candidate tokens is then computed from the overlap between their Wikipedia neighborhoods (see \Cref{eq:non_contradiction}). 

A potential limitation of this heuristic based on Wikipedia tokenization is that semantically related concepts, such as synonyms, hypernyms, or entities from the same category, may also have overlapping Wikipedia neighborhoods, resulting in an inappropriately low non-contradiction score. To test whether the heuristic captures non-contradiction rather than semantic similarity, we prompted a language model to construct three sets of 100 word pairs: (i) synonyms, e.g., \textit{car--automobile}, (ii) hypernyms, e.g., \textit{dog--animal}, and (iii) semantically close but non-hypernym concepts, e.g., \textit{France--Germany}. The latter set covers ten categories outlined in \Cref{tab:wikipedia_heuristic}. The full list of words is available in our repository at \url{https://github.com/orailix/kernel_token_contradiction}.

Our heuristic was designed to assign a low score to contradicting pairs of words where a substitution between the two words in a sentence would result in two contradicting statements. Therefore, we want our heuristic to assign low scores to semantically close, non-hypernym pairs, because words such as \textit{France} and \textit{Germany} are not substitutable. On the other hand, we want it to assign high scores to synonyms and hypernyms, because substituting a word like \textit{car} with \textit{automobile} or \textit{vehicle} does not contradict the original sentence containing \textit{car}. Our results in \Cref{tab:wikipedia_heuristic} confirm that this behavior is observed in practice, which supports the interpretation of our heuristic as a measure of substitutability rather than generic semantic similarity. Some categories, such as animals and space objects, nevertheless exhibit relatively high scores, reflecting the limitations of a corpus-based heuristic. However, KTC does not use these pairwise scores alone. Steps [B], [C], and [D], discussed in \Cref{sec:methodology}, aggregate the relationships between multiple candidate tokens, as well as their predictive distributions, resulting in a measure that is much more robust than individual token-pair scores (see empirical evaluations in \Cref{sec:empirical_results}).

\begin{table}[ht!]
\centering
\caption[Non-contradiction scores produced by the Wikipedia heuristic]{\textbf{Non-contradiction scores produced by the Wikipedia heuristic.} Synonyms and hypernyms receive substantially higher scores than semantically close but non-hypernym concepts, indicating that the heuristic does not simply measure generic semantic similarity. Values are mean $\pm$ standard deviation over the pairs of the category.}
\label{tab:wikipedia_heuristic}
\begin{tabular}{lc}
    \toprule
    \textbf{Pair type / category} & \textbf{Non-contradiction} \\
    \midrule
    Synonyms & $0.860 \pm 0.152$ \\
    Hypernyms & $0.837 \pm 0.132$ \\
    Semantically close, non-hypernym & $0.574 \pm 0.234$ \\
    \midrule
    European countries & $0.475 \pm 0.146$ \\
    Social media platforms & $0.525 \pm 0.146$ \\
    Months & $0.238 \pm 0.153$ \\
    Computer hardware & $0.763 \pm 0.205$ \\
    Animals & $0.825 \pm 0.100$ \\
    French cities & $0.538 \pm 0.202$ \\
    Car models & $0.475 \pm 0.094$ \\
    Programming languages & $0.538 \pm 0.217$ \\
    Space objects & $0.750 \pm 0.209$ \\
    Cocktails & $0.613 \pm 0.142$ \\
    \bottomrule
\end{tabular}
\end{table}

\section{ROC Curves} \label{app:roc_curves}

\begin{figure}[hb!]
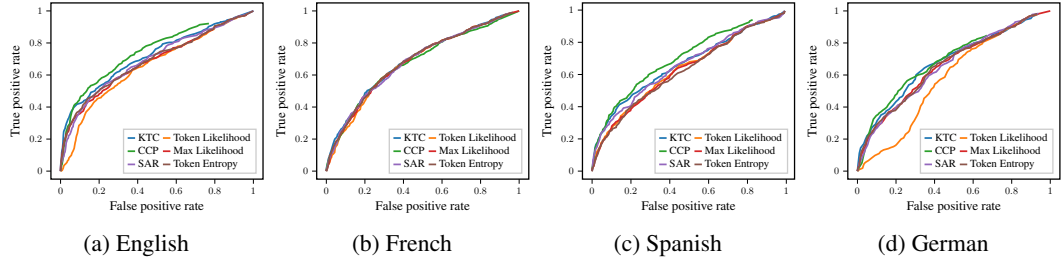

	\centering
	\hfill
	\begin{subfigure}{.245\textwidth}
		\centering
		\small
		\resizebox{\textwidth}{!}{\input{figures/results/roc_per_lang/roc_per_lang_0_0.tikz}}
		\caption{English}
	\end{subfigure}
	\hfill
	\begin{subfigure}{.245\textwidth}
		\centering
		\small
		\resizebox{\textwidth}{!}{\input{figures/results/roc_per_lang/roc_per_lang_1_0.tikz}}
		\caption{French}
	\end{subfigure}
	\hfill
	\begin{subfigure}{.245\textwidth}
		\centering
		\small
		\resizebox{\textwidth}{!}{\input{figures/results/roc_per_lang/roc_per_lang_0_1.tikz}}
		\caption{Spanish}
	\end{subfigure}
	\begin{subfigure}{.245\textwidth}
		\centering
		\small
		\resizebox{\textwidth}{!}{\input{figures/results/roc_per_lang/roc_per_lang_1_1.tikz}}
		\caption{German}
	\end{subfigure}
	\hfill
	\caption{\textbf{ROC curves for claim-level hallucination detection} across four languages, using product for token-to-claim aggregation. Results are averaged over 16 LLMs across two benchmarks. Each method is evaluated over its full grid search space, and we report its best-performing configuration.}\label{fig:roc_per_lang}
\end{figure}

\begin{figure}[ht!]
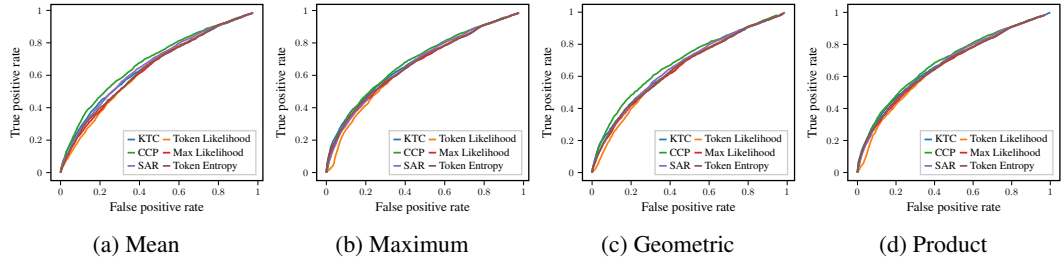

	\centering
	\hfill
	\begin{subfigure}{.245\textwidth}
		\centering
		\small
		\resizebox{\textwidth}{!}{\input{figures/results/roc_per_agg/roc_per_agg_0_0.tikz}}
		\caption{Mean}
	\end{subfigure}
	\hfill
	\begin{subfigure}{.245\textwidth}
		\centering
		\small
		\resizebox{\textwidth}{!}{\input{figures/results/roc_per_agg/roc_per_agg_1_0.tikz}}
		\caption{Maximum}
	\end{subfigure}
	\hfill
	\begin{subfigure}{.245\textwidth}
		\centering
		\small
		\resizebox{\textwidth}{!}{\input{figures/results/roc_per_agg/roc_per_agg_0_1.tikz}}
		\caption{Geometric}
	\end{subfigure}
	\begin{subfigure}{.245\textwidth}
		\centering
		\small
		\resizebox{\textwidth}{!}{\input{figures/results/roc_per_agg/roc_per_agg_1_1.tikz}}
		\caption{Product}
	\end{subfigure}
	\hfill

    \caption{\textbf{ROC curves under four token-to-claim aggregation strategies} (Mean, Maximum, Geometric mean, Product), averaged over all four languages and 16 LLMs. Each method is evaluated over its full grid search space, and we report its best-performing configuration.}\label{fig:roc_per_agg}
\end{figure}

In addition to the Precision-Recall curves outlined in \Cref{sec:empirical_results}, we report the ROC curves for the top-performing hyperparameter configurations of each method. \Cref{fig:roc_per_lang} presents the top-performing results in each language (English, French, Spanish, and German), while \Cref{fig:roc_per_agg} presents the top-performing results depending on the token-to-claim aggregation strategy. KTC achieves the second-best performance, closely following CCP.

\section{Hyperparameter Tuning and Generalization} \label{app:hyperparameter_tuning}

We evaluate all methods over predefined hyperparameter grids, summarized in \Cref{tab:hyperparameter_grid}. For KTC, this gives 30 configurations. We select the seven best configurations according to ROC-AUC and the seven best according to PR-AUC; after removing overlaps, this results in the eight configurations reported in the main paper.

For SAR and CCP, similar to what is done for the MUCH benchmark~\citep{dentan_much_2026}, we introduce a parameter $\sigma$ that quantifies the number of preceding and succeeding tokens that are fed to the NLI model to compute the contradiction, entailment, or neutral relationship. We also introduce a Small variant using a 70M NLI model even though it was not considered in the original papers~\citep{fadeeva_factchecking_2024,duan_shifting_2024}, to enable faster inference and assess whether this could improve their accuracy--efficiency trade-off relative to KTC. Our experiments presented in \Cref{sec:empirical_results} show that it does not yield a better trade-off than KTC. Overall, we evaluate 9 and 18 hyperparameter configurations for SAR and CCP, respectively.

\begin{itemize}
    \item Small (70M parameters): \texttt{MoritzLaurer/DeBERTa-v3-xsmall-mnli-fever-anli-ling-binary}
    \item Medium (100M parameters): \texttt{microsoft/deberta-base-mnli}
    \item Large (350M parameters): \texttt{microsoft/deberta-large-mnli}
\end{itemize}

\begin{table}[ht!]
\centering
\caption{Hyperparameter configurations evaluated for each method.}
\label{tab:hyperparameter_grid}
\begin{tabular}{lll}
\toprule
\textbf{Method} & \textbf{Hyperparameter} & \textbf{Values} \\
\midrule
KTC
& $\nu$ & $\{3,4,5,6,8\}$ \\
& $\tau$ & $\{0.1,0.2,0.3,0.4,0.5,0.7\}$ \\
& $\delta$ & $24$ (fixed) \\
\midrule
SAR
& $\sigma$ & $\{3,5,8\}$ \\
& NLI model & Small, Medium, Large \\
\midrule
CCP
& $\sigma$ & $\{3,5,8\}$ \\
& $\delta$ & $\{10,24\}$ \\
& NLI model & Small, Medium, Large \\
\midrule
Token Entropy
& $\delta$ & $\{5,10,24\}$ \\
\midrule
Likelihood
& Variant & Token, Max \\
\bottomrule
\end{tabular}
\end{table}

\subsection{Evaluation on the \textit{Main} Dataset}

\Cref{fig:compute_time} reports the performance of the 8 best KTC configurations, as well as the performance of all configurations for the baselines. Time is reported relative to LLM generation time. For CCP and SAR, the same predictions are obtained on CPU and GPU; only the computation time differs. Detailed numerical values are provided in \Cref{tab:compute_time_table}.

\begingroup
\begin{table}[!t]
    \centering
    \caption[Claim-level UQ performance on our main evaluation set]{\textbf{Performance of the evaluated hyperparameter configurations on our \textit{main} evaluation set} (698 manually-annotated samples from MUCH~\citep{dentan_much_2026} and Mu-SHROOM~\citep{vazquez_semeval2025_2025}). Time is relative to LLM generation time.}
    \begin{tabular}{llcccc}
        \toprule
        \textbf{Method} & \textbf{Configuration} & \textbf{GPU time} & \textbf{CPU time} & \textbf{ROC-AUC} & \textbf{PR-AUC} \\
        \midrule
        KTC & $\nu=3,\tau=0.2$ & -- & 0.0566 & 0.6795 & 0.7483 \\
        KTC & $\nu=3,\tau=0.3$ & -- & 0.0566 & 0.6786 & 0.7590 \\
        KTC & $\nu=4,\tau=0.2$ & -- & 0.0566 & 0.6787 & 0.7540 \\
        KTC & $\nu=4,\tau=0.3$ & -- & 0.0566 & 0.6779 & 0.7808 \\
        KTC & $\nu=4,\tau=0.4$ & -- & 0.0566 & 0.6725 & 0.7704 \\
        KTC & $\nu=5,\tau=0.2$ & -- & 0.0566 & 0.6767 & 0.7515 \\
        KTC & $\nu=5,\tau=0.3$ & -- & 0.0566 & 0.6742 & 0.7794 \\
        KTC & $\nu=6,\tau=0.2$ & -- & 0.0566 & 0.6751 & 0.7434 \\
        \midrule
        CCP & $\delta=10,\sigma=3$, Large & 0.8812 & 11.4870 & 0.6809 & 0.5923 \\
        CCP & $\delta=10,\sigma=3$, Medium & 0.4611 & 3.6900 & 0.6670 & 0.6316 \\
        CCP & $\delta=10,\sigma=3$, Small & 0.3981 & 1.8883 & 0.5644 & 0.3449 \\
        CCP & $\delta=10,\sigma=5$, Large & 1.0260 & 14.6420 & 0.6885 & 0.6215 \\
        CCP & $\delta=10,\sigma=5$, Medium & 0.4925 & 4.5712 & 0.6616 & 0.6555 \\
        CCP & $\delta=10,\sigma=5$, Small & 0.4036 & 2.2030 & 0.5627 & 0.3420 \\
        CCP & $\delta=10,\sigma=8$, Large & 1.1440 & 18.4894 & 0.6859 & 0.6469 \\
        CCP & $\delta=10,\sigma=8$, Medium & 0.5445 & 5.8694 & 0.6437 & 0.6762 \\
        CCP & $\delta=10,\sigma=8$, Small & 0.4076 & 2.6515 & 0.5625 & 0.3463 \\
        CCP & $\delta=24,\sigma=3$, Large & 1.1589 & 20.6294 & 0.6828 & 0.6067 \\
        CCP & $\delta=24,\sigma=3$, Medium & 0.5500 & 6.5696 & 0.6730 & 0.6468 \\
        CCP & $\delta=24,\sigma=3$, Small & 0.4327 & 3.7530 & 0.5709 & 0.3493 \\
        CCP & $\delta=24,\sigma=5$, Large & 1.3470 & 27.5059 & 0.6913 & 0.6334 \\
        CCP & $\delta=24,\sigma=5$, Medium & 0.6184 & 8.5995 & 0.6703 & 0.6696 \\
        CCP & $\delta=24,\sigma=5$, Small & 0.4477 & 4.8387 & 0.5686 & 0.3472 \\
        CCP & $\delta=24,\sigma=8$, Large & 1.6263 & 37.3485 & 0.6900 & 0.6562 \\
        CCP & $\delta=24,\sigma=8$, Medium & 0.7207 & 11.6444 & 0.6500 & 0.6857 \\
        CCP & $\delta=24,\sigma=8$, Small & 0.4862 & 7.1833 & 0.5686 & 0.3516 \\
        \midrule
        SAR & $\sigma=3$, Large & 0.1109 & 2.1637 & 0.6721 & 0.6391 \\
        SAR & $\sigma=3$, Medium & 0.0637 & 0.7946 & 0.6673 & 0.6449 \\
        SAR & $\sigma=3$, Small & 0.0551 & 0.6137 & 0.6215 & 0.4420 \\
        SAR & $\sigma=5$, Large & 0.1314 & 2.9426 & 0.6765 & 0.6506 \\
        SAR & $\sigma=5$, Medium & 0.0708 & 1.0307 & 0.6640 & 0.6578 \\
        SAR & $\sigma=5$, Small & 0.0574 & 0.6924 & 0.6161 & 0.4286 \\
        SAR & $\sigma=8$, Large & 0.1629 & 4.0991 & 0.6735 & 0.6451 \\
        SAR & $\sigma=8$, Medium & 0.0818 & 1.3454 & 0.6616 & 0.6715 \\
        SAR & $\sigma=8$, Small & 0.0590 & 0.8183 & 0.6158 & 0.4230 \\
        \midrule
        Token Entropy & $\delta=5$ & -- & 0.0279 & 0.6635 & 0.6712 \\
        Token Entropy & $\delta=10$ & -- & 0.0279 & 0.6698 & 0.7040 \\
        Token Entropy & $\delta=24$ & -- & 0.0277 & 0.6730 & 0.6956 \\
        \midrule
        Likelihood & Token & -- & 0.0264 & 0.6608 & 0.4501 \\
        Likelihood & Max & -- & 0.0263 & 0.6692 & 0.6851 \\
        \bottomrule
    \end{tabular}
    \label{tab:compute_time_table}
\end{table}
\endgroup

\FloatBarrier

\subsection{Generalization to the \textit{Validation} Dataset}

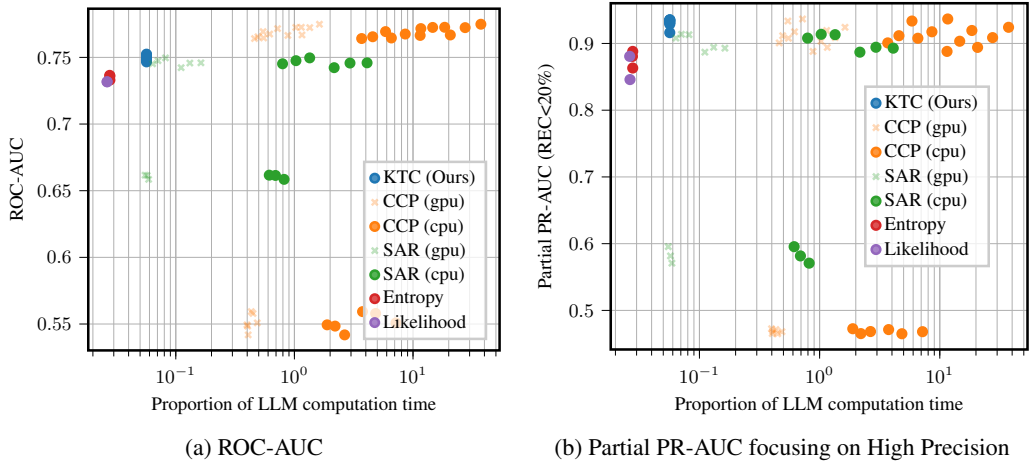
\begin{figure}[h!]
	\hfill
	\begin{subfigure}[t]{.49\textwidth}
		\small
		\centering
		\resizebox{.97\textwidth}{!}{
\begin{tikzpicture}

\definecolor{crimson2143940}{RGB}{214,39,40}
\definecolor{darkgray176}{RGB}{176,176,176}
\definecolor{darkorange25512714}{RGB}{255,127,14}
\definecolor{forestgreen4416044}{RGB}{44,160,44}
\definecolor{lightgray204}{RGB}{204,204,204}
\definecolor{mediumpurple148103189}{RGB}{148,103,189}
\definecolor{steelblue31119180}{RGB}{31,119,180}

\begin{groupplot}[group style={group size=2 by 1}]
\nextgroupplot[
every mark/.append style={solid}, very thick,
legend cell align={left},
legend style={
  fill opacity=0.8,
  draw opacity=1,
  text opacity=1,
  at={(0.97,0.03)},
  anchor=south east,
  draw=lightgray204
},
log basis x={10},
tick align=outside,
tick pos=left,
x grid style={darkgray176},
xlabel={Proportion of LLM computation time},
xmajorgrids,
xmin=0.0182796248282595, xmax=53.6917203356727,
xminorgrids,
xmode=log,
xtick style={color=black},
y grid style={darkgray176},
ylabel={ROC-AUC},
ymajorgrids,
ymin=0.530201226133871, ymax=0.786521020204657,
yminorgrids,
ytick style={color=black}
]
\addplot [draw=steelblue31119180, fill=steelblue31119180, mark=*, only marks]
table{%
x  y
0.0566483084185681 0.748028740078772
0.0566483084185681 0.750256056246547
0.0566483084185681 0.748165844943458
0.0566483084185681 0.752447583151117
0.0566483084185681 0.748122154554815
0.0566483084185681 0.747620328316291
0.0566483084185681 0.751986260965939
0.0566483084185681 0.746611572772268
};
\addlegendentry{KTC (Ours)}
\addplot [draw=darkorange25512714, fill=darkorange25512714, mark=x, only marks, opacity=0.3]
table{%
x  y
0.881195908733281 0.766475951248355
0.461054287962234 0.764140650978532
0.398111723052714 0.549383593267418
1.02596380802518 0.772472498690737
0.492525570416995 0.765477882561986
0.403619197482297 0.548435908122755
1.14398111723053 0.772580149984069
0.544453186467349 0.769342956583636
0.407553107789142 0.541852125864361
1.15892997639654 0.766822811887775
0.549960660896932 0.764553000977484
0.43273013375295 0.559305878364602
1.34697088906373 0.772327252338761
0.618410700236035 0.767506758451366
0.447678992918961 0.557914454428583
1.62627852084972 0.774870120474167
0.720692368214005 0.771603220814281
0.486231313926042 0.550852635591725
};
\addlegendentry{CCP (gpu)}
\addplot [draw=darkorange25512714, fill=darkorange25512714, mark=*, only marks]
table{%
x  y
11.4870180959874 0.766475951248355
3.69000786782061 0.764140650978532
1.8882769472856 0.549383593267418
14.6420141620771 0.772472498690737
4.57120377655389 0.765477882561986
2.2029897718332 0.548435908122755
18.4893784421715 0.772580149984069
5.86939417781275 0.769342956583636
2.65145554681353 0.541852125864361
20.6294256490952 0.766822811887775
6.56963021243116 0.764553000977484
3.75295043273013 0.559305878364602
27.5059008654603 0.772327252338761
8.59952793076318 0.767506758451366
4.83870967741935 0.557914454428583
37.3485444531865 0.774870120474167
11.6443745082612 0.771603220814281
7.18332022029898 0.550852635591725
};
\addlegendentry{CCP (cpu)}
\addplot [draw=forestgreen4416044, fill=forestgreen4416044, mark=x, only marks, opacity=0.3]
table{%
x  y
0.110936270653029 0.74228883129819
0.0637293469708891 0.745203113960496
0.0550747442958301 0.661659029507844
0.131392604248623 0.745745173074892
0.0708103855232101 0.747538186194717
0.0574350904799371 0.66144138493392
0.162863886703383 0.745924322774015
0.0818253343823761 0.749624699623185
0.0590086546026751 0.658512495666861
};
\addlegendentry{SAR (gpu)}
\addplot [draw=forestgreen4416044, fill=forestgreen4416044, mark=*, only marks]
table{%
x  y
2.16365066876475 0.74228883129819
0.794649881982691 0.745203113960496
0.613690007867821 0.661659029507844
2.94256490952006 0.745745173074892
1.03068450039339 0.747538186194717
0.692368214004721 0.66144138493392
4.09913453973249 0.745924322774015
1.34539732494099 0.749624699623185
0.818253343823761 0.658512495666861
};
\addlegendentry{SAR (cpu)}
\addplot [draw=crimson2143940, fill=crimson2143940, mark=*, only marks]
table{%
x  y
0.0278520849724626 0.733031806171514
0.0278520849724626 0.736058935640994
0.0276947285601888 0.73654621690601
};
\addlegendentry{Entropy}
\addplot [draw=mediumpurple148103189, fill=mediumpurple148103189, mark=*, only marks]
table{%
x  y
0.0264358772619984 0.731703036557817
0.0262785208497246 0.731691721061478
};
\addlegendentry{Likelihood}
\end{groupplot}

\end{tikzpicture}}
		\caption{ROC-AUC}
	\end{subfigure}
	\hfill
	\begin{subfigure}[t]{.49\textwidth}
		\small
		\centering
		\resizebox{.97\textwidth}{!}{
\begin{tikzpicture}

\definecolor{crimson2143940}{RGB}{214,39,40}
\definecolor{darkgray176}{RGB}{176,176,176}
\definecolor{darkorange25512714}{RGB}{255,127,14}
\definecolor{forestgreen4416044}{RGB}{44,160,44}
\definecolor{lightgray204}{RGB}{204,204,204}
\definecolor{mediumpurple148103189}{RGB}{148,103,189}
\definecolor{steelblue31119180}{RGB}{31,119,180}

\begin{groupplot}[group style={group size=2 by 1}]
\nextgroupplot[
every mark/.append style={solid}, very thick,
legend cell align={left},
legend style={
  fill opacity=0.8,
  draw opacity=1,
  text opacity=1,
  at={(0.91,0.5)},
  anchor=east,
  draw=lightgray204
},
log basis x={10},
tick align=outside,
tick pos=left,
x grid style={darkgray176},
xlabel={Proportion of LLM computation time},
xmajorgrids,
xmin=0.0182796248282595, xmax=53.6917203356727,
xminorgrids,
xmode=log,
xtick style={color=black},
y grid style={darkgray176},
ylabel={Partial PR-AUC (REC<20\%)},
ymajorgrids,
ymin=0.441455615256941, ymax=0.960241989265076,
yminorgrids,
ytick style={color=black}
]
\addplot [draw=steelblue31119180, fill=steelblue31119180, mark=*, only marks]
table{%
x  y
0.0566483084185681 0.93542140983756
0.0566483084185681 0.931111159266106
0.0566483084185681 0.934865261347727
0.0566483084185681 0.935216090077582
0.0566483084185681 0.91619610688901
0.0566483084185681 0.931105361332845
0.0566483084185681 0.933583023018994
0.0566483084185681 0.928908702958201
};
\addlegendentry{KTC (Ours)}
\addplot [draw=darkorange25512714, fill=darkorange25512714, mark=x, only marks, opacity=0.3]
table{%
x  y
0.881195908733281 0.888081504087218
0.461054287962234 0.900813998960471
0.398111723052714 0.472538150899212
1.02596380802518 0.903308688324015
0.492525570416995 0.911587347129908
0.403619197482297 0.465248236691664
1.14398111723053 0.919410171135353
0.544453186467349 0.933569476284178
0.407553107789142 0.468333735560139
1.15892997639654 0.894087851727675
0.549960660896932 0.907784961799961
0.43273013375295 0.471267222980899
1.34697088906373 0.908735153044434
0.618410700236035 0.917486995699454
0.447678992918961 0.465036814075492
1.62627852084972 0.924195992390257
0.720692368214005 0.936660790446525
0.486231313926042 0.468123681530285
};
\addlegendentry{CCP (gpu)}
\addplot [draw=darkorange25512714, fill=darkorange25512714, mark=*, only marks]
table{%
x  y
11.4870180959874 0.888081504087218
3.69000786782061 0.900813998960471
1.8882769472856 0.472538150899212
14.6420141620771 0.903308688324015
4.57120377655389 0.911587347129908
2.2029897718332 0.465248236691664
18.4893784421715 0.919410171135353
5.86939417781275 0.933569476284178
2.65145554681353 0.468333735560139
20.6294256490952 0.894087851727675
6.56963021243116 0.907784961799961
3.75295043273013 0.471267222980899
27.5059008654603 0.908735153044434
8.59952793076318 0.917486995699454
4.83870967741935 0.465036814075492
37.3485444531865 0.924195992390257
11.6443745082612 0.936660790446525
7.18332022029898 0.468123681530285
};
\addlegendentry{CCP (cpu)}
\addplot [draw=forestgreen4416044, fill=forestgreen4416044, mark=x, only marks, opacity=0.3]
table{%
x  y
0.110936270653029 0.886935927565744
0.0637293469708891 0.908021893222145
0.0550747442958301 0.595506285888614
0.131392604248623 0.894305746035241
0.0708103855232101 0.913693613157787
0.0574350904799371 0.581654089344366
0.162863886703383 0.892941215765274
0.0818253343823761 0.91315859642121
0.0590086546026751 0.570811481446721
};
\addlegendentry{SAR (gpu)}
\addplot [draw=forestgreen4416044, fill=forestgreen4416044, mark=*, only marks]
table{%
x  y
2.16365066876475 0.886935927565744
0.794649881982691 0.908021893222145
0.613690007867821 0.595506285888614
2.94256490952006 0.894305746035241
1.03068450039339 0.913693613157787
0.692368214004721 0.581654089344366
4.09913453973249 0.892941215765274
1.34539732494099 0.91315859642121
0.818253343823761 0.570811481446721
};
\addlegendentry{SAR (cpu)}
\addplot [draw=crimson2143940, fill=crimson2143940, mark=*, only marks]
table{%
x  y
0.0278520849724626 0.86318323505767
0.0278520849724626 0.888076702743574
0.0276947285601888 0.880581136191167
};
\addlegendentry{Entropy}
\addplot [draw=mediumpurple148103189, fill=mediumpurple148103189, mark=*, only marks]
table{%
x  y
0.0264358772619984 0.845879834806006
0.0262785208497246 0.880678122422914
};
\addlegendentry{Likelihood}
\end{groupplot}

\end{tikzpicture}}
		\caption{PR-AUC focusing on High Precision}
	\end{subfigure}
	\hfill
    \caption[Hyperparameter generalization of KTC]{\textbf{Hyperparameter generalization.} Each point corresponds to a method evaluated under a different set of hyperparameters. The $x$-axis reports the UQ computation time as a fraction of the LLM generation cost. KTC configurations are selected on the manually annotated data and evaluated without retuning, on our validation set, showing that the accuracy-efficiency trade-off generalizes to unseen data. See detailed numbers in \Cref{tab:compute_time_test_much_table}.}
    \label{fig:compute_time_test_much}
\end{figure}

To assess whether the selected KTC hyperparameters generalize beyond the manually annotated data, we evaluate the eight selected configurations without retuning, on our \textit{validation} set introduced in \Cref{sec:empirical_results} and consisting of the 4,673 automatically annotated samples from MUCH~\citep{dentan_much_2026}, which are excluded from the experiments on the \textit{main} set. We compare them with all configurations of SAR and CCP. Results are reported in \Cref{fig:compute_time_test_much} and \Cref{tab:compute_time_test_much_table}. The selected KTC configurations retain a competitive trade-off between performance and computational efficiency, with ROC-AUC ranging from $0.7466$ to $0.7524$ and PR-AUC from $0.9162$ to $0.9354$. Moreover, CCP and KTC are on par in the high-precision regime (PR-AUC at recall $\leq 20\%$, with CCP at $0.9367$ and KTC at $0.9354$) on this \textit{validation} set, whereas KTC outperforms CCP in this regime on the \textit{main} evaluation set. However, the noise induced by the automated annotations of the \textit{validation} set, compared to the human annotations of the \textit{main} evaluation set, weakens these results, especially in the high-precision regime, where a few incorrect annotations can heavily impact the evaluation.

Overall, these results on the \textit{validation} set show that KTC retains its competitive trade-off between performance and computational efficiency, as well as its strong results in the high-precision regime. This supports the generalization of the selected hyperparameters at larger scale, beyond the \textit{main} evaluation set used for hyperparameter tuning.

\begingroup
\begin{table}[!p]
    \centering
    \caption[Claim-level UQ performance on our validation set]{\textbf{Performance of the evaluated hyperparameter configurations on our \textit{validation} set} (4,673 automatically annotated samples from MUCH~\citep{dentan_much_2026}). Time is relative to LLM generation time. The KTC configurations are evaluated without retuning.}
    \begin{tabular}{llcccc}
        \toprule
        \textbf{Method} & \textbf{Configuration} & \textbf{GPU time} & \textbf{CPU time} & \textbf{ROC-AUC} & \textbf{PR-AUC} \\
        \midrule
        KTC & $\nu=3,\tau=0.2$ & -- & 0.0566 & 0.7480 & 0.9354 \\
        KTC & $\nu=3,\tau=0.3$ & -- & 0.0566 & 0.7503 & 0.9311 \\
        KTC & $\nu=4,\tau=0.2$ & -- & 0.0566 & 0.7482 & 0.9349 \\
        KTC & $\nu=4,\tau=0.3$ & -- & 0.0566 & 0.7524 & 0.9352 \\
        KTC & $\nu=4,\tau=0.4$ & -- & 0.0566 & 0.7481 & 0.9162 \\
        KTC & $\nu=5,\tau=0.2$ & -- & 0.0566 & 0.7476 & 0.9311 \\
        KTC & $\nu=5,\tau=0.3$ & -- & 0.0566 & 0.7520 & 0.9336 \\
        KTC & $\nu=6,\tau=0.2$ & -- & 0.0566 & 0.7466 & 0.9289 \\
        \midrule
        CCP & $\delta=10,\sigma=3$, Large & 0.8812 & 11.4870 & 0.7665 & 0.8881 \\
        CCP & $\delta=10,\sigma=3$, Medium & 0.4611 & 3.6900 & 0.7641 & 0.9008 \\
        CCP & $\delta=10,\sigma=3$, Small & 0.3981 & 1.8883 & 0.5494 & 0.4725 \\
        CCP & $\delta=10,\sigma=5$, Large & 1.0260 & 14.6420 & 0.7725 & 0.9033 \\
        CCP & $\delta=10,\sigma=5$, Medium & 0.4925 & 4.5712 & 0.7655 & 0.9116 \\
        CCP & $\delta=10,\sigma=5$, Small & 0.4036 & 2.2030 & 0.5484 & 0.4652 \\
        CCP & $\delta=10,\sigma=8$, Large & 1.1440 & 18.4894 & 0.7726 & 0.9194 \\
        CCP & $\delta=10,\sigma=8$, Medium & 0.5445 & 5.8694 & 0.7693 & 0.9336 \\
        CCP & $\delta=10,\sigma=8$, Small & 0.4076 & 2.6515 & 0.5419 & 0.4683 \\
        CCP & $\delta=24,\sigma=3$, Large & 1.1589 & 20.6294 & 0.7668 & 0.8941 \\
        CCP & $\delta=24,\sigma=3$, Medium & 0.5500 & 6.5696 & 0.7646 & 0.9078 \\
        CCP & $\delta=24,\sigma=3$, Small & 0.4327 & 3.7530 & 0.5593 & 0.4713 \\
        CCP & $\delta=24,\sigma=5$, Large & 1.3470 & 27.5059 & 0.7723 & 0.9087 \\
        CCP & $\delta=24,\sigma=5$, Medium & 0.6184 & 8.5995 & 0.7675 & 0.9175 \\
        CCP & $\delta=24,\sigma=5$, Small & 0.4477 & 4.8387 & 0.5579 & 0.4650 \\
        CCP & $\delta=24,\sigma=8$, Large & 1.6263 & 37.3485 & 0.7749 & 0.9242 \\
        CCP & $\delta=24,\sigma=8$, Medium & 0.7207 & 11.6444 & 0.7716 & 0.9367 \\
        CCP & $\delta=24,\sigma=8$, Small & 0.4862 & 7.1833 & 0.5509 & 0.4681 \\
        \midrule
        SAR & $\sigma=3$, Large & 0.1109 & 2.1637 & 0.7423 & 0.8869 \\
        SAR & $\sigma=3$, Medium & 0.0637 & 0.7946 & 0.7452 & 0.9080 \\
        SAR & $\sigma=3$, Small & 0.0551 & 0.6137 & 0.6617 & 0.5955 \\
        SAR & $\sigma=5$, Large & 0.1314 & 2.9426 & 0.7457 & 0.8943 \\
        SAR & $\sigma=5$, Medium & 0.0708 & 1.0307 & 0.7475 & 0.9137 \\
        SAR & $\sigma=5$, Small & 0.0574 & 0.6924 & 0.6614 & 0.5817 \\
        SAR & $\sigma=8$, Large & 0.1629 & 4.0991 & 0.7459 & 0.8929 \\
        SAR & $\sigma=8$, Medium & 0.0818 & 1.3454 & 0.7496 & 0.9132 \\
        SAR & $\sigma=8$, Small & 0.0590 & 0.8183 & 0.6585 & 0.5708 \\
        \midrule
        Token Entropy & $\delta=5$ & -- & 0.0279 & 0.7330 & 0.8632 \\
        Token Entropy & $\delta=10$ & -- & 0.0279 & 0.7361 & 0.8881 \\
        Token Entropy & $\delta=24$ & -- & 0.0277 & 0.7365 & 0.8806 \\
        \midrule
        Likelihood & Token & -- & 0.0264 & 0.7317 & 0.8459 \\
        Likelihood & Max & -- & 0.0263 & 0.7317 & 0.8807 \\
        \bottomrule
    \end{tabular}
    \label{tab:compute_time_test_much_table}
\end{table}
\endgroup

\section{Reproducible Science} \label{app:reproducible_science}

\paragraph{Data and Code Availability}

The datasets used in this paper are publicly available. The MUCH dataset is available at \url{https://huggingface.co/datasets/orailix/MUCH}, and the Mu-SHROOM dataset is available at \url{https://huggingface.co/datasets/Helsinki-NLP/mu-shroom}. Alongside this paper, we release a code repository to reproduce our results.

\paragraph{Licenses} In this paper, we use two datasets: MUCH and Mu-SHROOM. They are properly cited in the main content of the paper. The former is released under the \href{https://choosealicense.com/licenses/apache-2.0/}{Apache License 2.0} and the latter is released under the \href{https://choosealicense.com/licenses/cc-by-4.0/}{Creative Commons Attribution 4.0 license}. Furthermore, NLI models are used in some of the baselines. We use the \texttt{microsoft/deberta-base-mnli}, \texttt{microsoft/deberta-large-mnli}, and \texttt{MoritzLaurer/DeBERTa-v3-xsmall-mnli-fever-anli-ling-binary} models, available through the \href{https://choosealicense.com/licenses/mit/}{MIT license}.

\paragraph{Reproducibility}

All the results are completely reproducible. We release a code repository that includes the necessary information on library versions to reproduce our environment and run all the experiments and reproduce the results. Additional details on hyperparameter tuning are provided in \Cref{app:hyperparameter_tuning}, and exact values are also included in the code repository.

\begin{center}
  \begin{minipage}{\linewidth}
    \centering
    \raisebox{-0.2\height}{\includegraphics[width=1em]{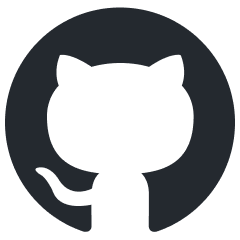}}%
    \quad
    {\small\texttt{\href{https://github.com/orailix/kernel_token_contradiction}{\tt orailix/kernel\_token\_contradiction}}}
  \end{minipage}
\end{center}

\end{document}